\documentclass[letterpaper, 10 pt, journal, twoside]{IEEEtran} 

\IEEEoverridecommandlockouts                              

\usepackage{booktabs}
\usepackage{multirow}
\usepackage{graphicx}
\usepackage{subfig}
\usepackage{booktabs}
\usepackage{color,soul}
\usepackage{setspace}
\usepackage{pbox}
\usepackage{booktabs}
\usepackage[final]{pdfpages}

\usepackage{xcolor}

\title{
Multi-Modal Trip Hazard Affordance Detection\\On Construction Sites
}

\author{Sean McMahon, Niko S\"underhauf, Ben Upcroft, and Michael Milford%
\thanks{Manuscript received: February 15, 2017; Revised May 16, 2017; Accepted June 5, 2017.}
\thanks{This paper was recommended for publication by Editor Wan Kyun Chung upon evaluation of the Associate Editor and Reviewers' comments.
This research was supported by an APA scholarship and by the Australian Research Council Centre
of Excellence for Robotic Vision (project number CE140100016).}
\thanks{The authors are with the ARC Centre of Excellence for Robotic Vision, Queensland University of Technology (QUT),
Brisbane QLD 4000, Australia. {\tt\small www.roboticvision.org}. Contact:~{\tt\small sean.mcmahon@roboticvision.org}}
}

\begin{document}\
\null%
\includepdf[pages=1]{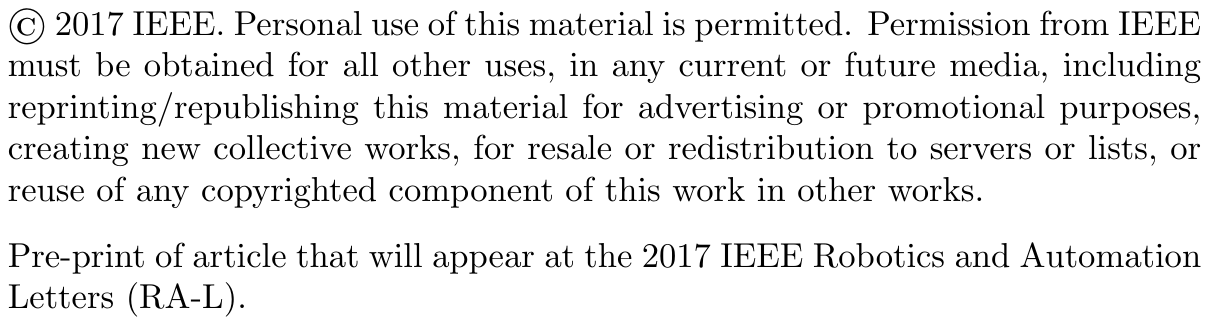}

\maketitle
\markboth{IEEE Robotics and Automation Letters. Preprint Version. Accepted June 2017}
{McMahon \MakeLowercase{\textit{et al.}}: Multi-Modal Trip Hazard Affordance Detection On Construction Sites}

\begin{abstract}

Trip hazards are a significant contributor to accidents on construction and manufacturing sites, where over a third of Australian workplace injuries occur \cite{key_WHS_2014}. Current safety inspections are labour intensive and limited by human fallibility, making automation of trip hazard detection appealing from both a safety and economic perspective.
Trip hazards present an interesting challenge to modern learning techniques because they are defined as much by affordance as by object type; for example wires on a table are not a trip hazard, but can be if lying on the ground.
To address these challenges, we conduct a comprehensive investigation into the performance characteristics of 11 different colour and depth fusion approaches, including 4 fusion and one non fusion approach; using colour and two types of depth images. Trained and tested on over 600 labelled trip hazards over 4 floors and 2000m\textsuperscript{2} in an active construction site, this approach was able to differentiate between identical objects in different physical configurations (see Figure 1). Outperforming a colour-only detector, our multi-modal trip detector fuses colour and depth information to achieve a 4\% absolute improvement in F1-score.
These investigative results and the extensive publicly available dataset moves us one step closer to assistive or fully automated safety inspection systems on construction sites.

\end{abstract}

\section{INTRODUCTION} \label{intro}

\IEEEPARstart{T}{rip} hazards are one of the most prevalent dangers on construction sites; for example on Australian construction sites 36 people are killed each year and 35 people are seriously injured everyday with a median cost of \$10'000 per injury \cite{SafeWorkAustralia2015}. Because detecting trip hazards is currently an entirely manual process performed by safety inspectors and workers, automating trip hazard detection is critical from both a safety and economic perspective.

Automating visual trip hazard detection however is not straightforward, whether in the form of an assistive or fully automated camera or robotic based solution. Construction sites are highly dynamic with different visual changes and hazards across different construction sites, with even the same location, visited over several days, appearing vastly different and presenting a new set of trip hazards. Deploying a conventional object-orientated approach is limited because trip hazards are an affordance-based concept, rather than object-based; a ladder leaning against a wall is less of a trip hazard than a ladder lying flat on the ground (Figure 1).

Therefore an investigation determining how to most effectively fuse information from colour and depth modalities to directly detect trip hazard affordances, is conducted. As a part of this investigation we evaluate the performance characteristics of a non fusion and 4 fusion approaches (see Figure~\ref{fig:all_fusion}). Each fusion approach combines colour images with depth or HHA images (see Section~\ref{section:networkTraining}). To execute the 4 different fusion approaches, a CNN adapted for semantic segmentation (dense classification) is used. Through the investigation of these colour and depth fusion techniques a 4\% absolute improvement in F1-score is achieved over the colour-only non fusion approach.

\begin{figure}[t]
\centering
\setlength\tabcolsep{1.0pt}
\begin{tabular}{cc}
\subfloat[Ground Truth]{\includegraphics[width=0.37\linewidth]{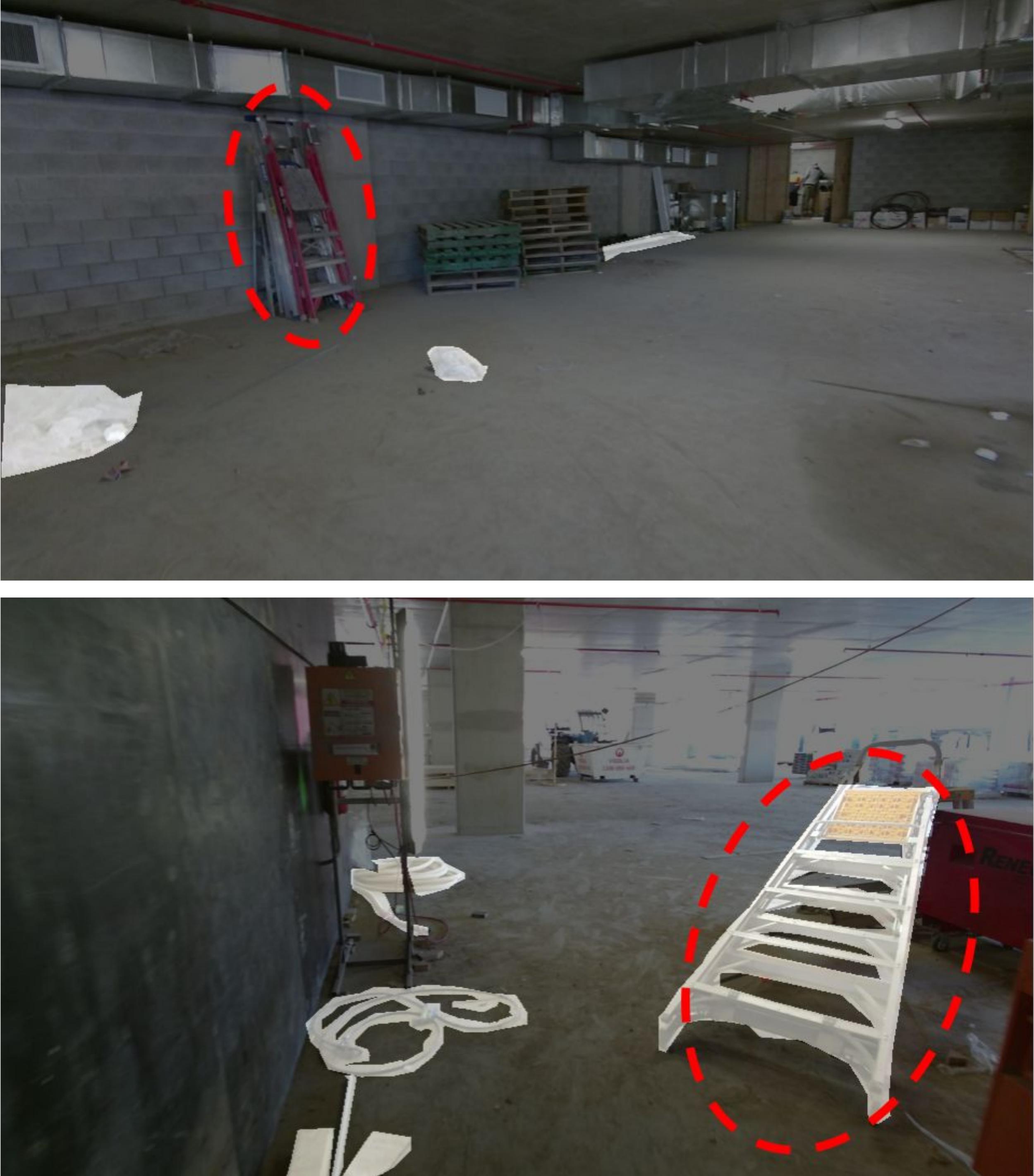}} &
\subfloat[Trip Hazard Predictions]{\includegraphics[width=0.37\linewidth]{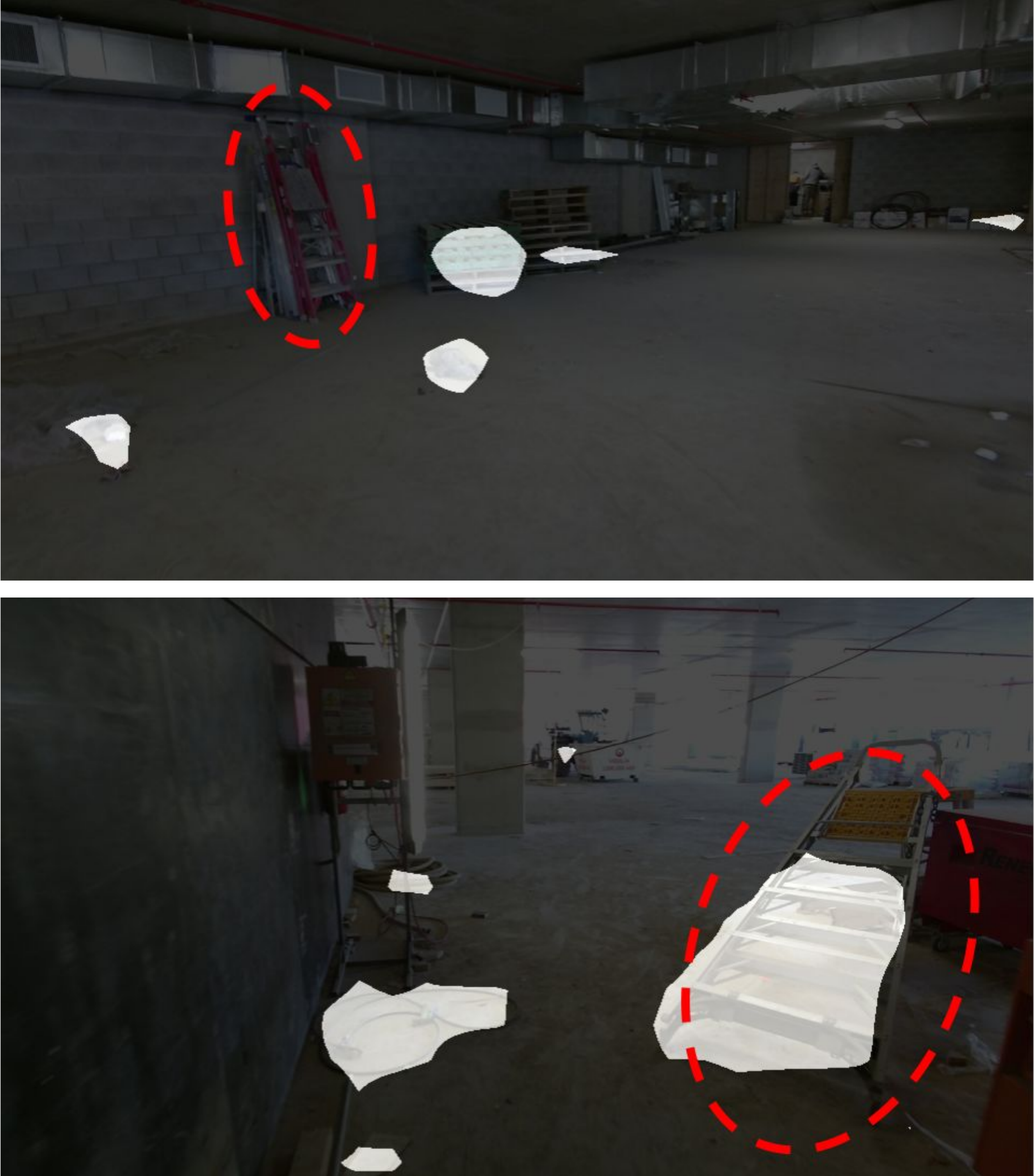}} \\
\end{tabular}
\caption{One of the proposed approaches fuses colour and depth information to correctly differentiate between the same object (ladder) as a trip hazard when lying down (2nd row) and as a non-trip hazard when standing up (1st row). The white regions show trip hazards.}
\label{fig:cover}
\vspace{-0.5cm}
\end{figure}

The key contributions of this research over previous research, including \cite{mcmah15}, are:
\begin{itemize}
\item Creation of a large scale construction site dataset covering 2000m\textsuperscript{2} of a construction site with over 600 labelled trip hazards, available at \textit{http://tinyurl.com/Tripdataset}.
\item An investigation into the performance characteristics of 11 different colour and depth fusion approaches, including no fusion, early fusion, mid fusion, late overlay fusion and late proportional fusion with a comprehensive evaluation of three different image types; colour, depth and HHA images.
\item Universal performance improvement from colour and depth fusion approach over the colour-only non fusion.
\item Improved performance in 4 of 5 metrics over an earlier prototype \cite{mcmah15} using an updated network with colour information only.

\end{itemize}

\section{Related Work}

This section presents related work categorised into the three areas, affordance detection, visual classification with Convolutional Neural Networks (CNNs) and Depth Information with CNNs. Trip hazards are a type of affordance, and this review of related work in affordance detection will cover learning affordances from interaction and visual affordance detection. As the trip detector uses a CNN adapted for semantic segmentation (dense classification), the second section reviews the research on CNNs trained with colour images. Because of our comprehensive investigation into colour and depth fusion, the final section reviews research on sensor fusion with CNNs.

\subsection{Affordance Detection}

Affordances were first introduced as a psychologist's term for the aspect of an object’s appearance that suggests what an agent can do to it, such as its liftability, squeezability  or even trippability \cite{Gibson}. As the action space (affordances) of objects can vary  widely between different robots, this work defines affordances within the human action space. Affordance detection has been divided into two main approaches, visual affordance  detection and learning affordances from interaction and geometric information.

\subsubsection{Learning Affordances from Interaction and Geometric Information}
There are two main approaches to learning affordances from interaction, through robotic interaction \cite{Goncalves2014, Mar2015} and human demonstration \cite{Chan2015, Koppula2015}​. The limitations of these techniques include the long time duration needed to interact intrusively with objects in the environment, which is impractical on construction sites. Additionally the robot or human must manipulate or interact with the environment, which is also impractical on construction sites both from a time and safety perspective; tripping over objects to learn the trip affordance would be impractical.

The use of scene geometry such as depth information is another popular approach in detecting affordances as it allows for direct observations of the shape of objects in a scene \cite{Myers2015, Pas2013-ec, Tunnermann2015}​.

\subsubsection{Visual Affordance Detection}
Visual affordance detection is the use of colour images to learn affordances without interaction and has been divided into three main approaches.
The first is direct affordance detection performed in this research, which involves learning affordances end-to-end directly from image pixels  \cite{Fouhey2015, mcmah15}. Second, indirect affordance detection is when known objects in a scene are linked to their affordances via an affordance knowledge tree or lookup table \cite{Chao2015, Chen2015b, Yao2013-cc}. The third approach is semi-direct affordance detection, which is the use of object attributes to detect object affordances \cite{Hassan2015, Moldovan2014}.

\subsection{Visual Classification with Convolutional Neural Networks (CNNs)}
Recently CNNs achieved current state-of-the-art performance for both image classification and object detection tasks \cite{He2015a, Ren2015a}. Image classification involves assigning an entire image one classification. AlexNet \cite{Krizhevsky2012} was the first CNN to achieve top results on image classification benchmarks \cite{ILSVRC15}, which was quickly superseded by VGG \cite{Simonyan2014a} and now ResNet \cite{He2015a}. The most common method for object detection introduced by \cite{Girshick2014} with R-CNN is to use a CNN to classify objects using image regions as inputs rather than the whole image. This has been further developed with Faster R-CNN \cite{Ren2015a} for top performance on object detection benchmarks \cite{ILSVRC15}. The task of dense classification, or semantic segmentation, is similar to that of object detection, but instead of a bounding box around objects, in dense classification every pixel of an image is assigned a prediction. Long et al. \cite{Long2014} performed this task by converting standard CNNs into a fully convolutional network (FCN). Since \cite{Long2014} there have been numerous other works which have improved performance in dense classification with similar approaches \cite{Chen2015c, Liu2015c, Liang2015a, Oliveira2016}.

\subsection{Depth Information with CNNs} \label{section:depthWithCNNs}
This section describes approaches used to train CNNs on colour and depth information.
Colour and depth based CNNs commonly incorporate the two modalities with a two channel approach, generally known as late fusion, where each channel has a set of convolution layers is trained on colour and depth information. These two channels are then fused together through a layer towards the end of a CNN, usually with additional convolutional layers on concatenated CNN features \cite{Long2014, Valada2016, Hou2016} or training a classifier on the CNN features \cite{Gupta2014a, Wang2016}. An alternative approach, early fusion, combines depth and colour information at the input of the CNN \cite{Long2014, Valada2016}. Long ​et al. \cite{Long2014}  and Valada et al. \cite{Valada2016} found the two channel late fusion approach produced better performance on their benchmark tests. Another interesting fusion approach \cite{Hazirbas2016} adopts a kind of network fusion where the features from colour and depth trained networks are summed together element-wise during testing. However, this approach is outperformed by an early fusion approach in the SUN-3D benchmark test set \cite{Hazirbas2016}. These comparisons give an indication that the late fusion approach might give higher detection rates, but for a definitive answer the depth incorporated CNNs must be tested on construction site data for trip hazard detection.

Apart from using different CNN architectures, Gupta et al. \cite{Gupta2014a} demonstrated that pre-processing depth information can yield improved performance over raw depth images. The depth pre-processing approach used was a three channel depth image called a HHA image (Horizontal disparity, Height above ground and the Angle the pixel normal makes with an inferred gravity direction) \cite{Gupta2013a}. The approaches discussed here are primarily trained and tested on the NYU dataset \cite{Silberman2012} (which has approximately 1100 colour and depth images). The depth information in this dataset is much denser than depth data collected on our construction site dataset.

\section{Approach}

This section establishes four CNN architectures used to investigate multi-modal trip hazard detection on construction sites. Subsection~\ref{section:classiferArch} details the 11 different fusion approaches investigated and the updated architecture of the trip detector over TripNet \cite{McMahon2015}. Section~\ref{section:networkTraining} will further explain how the 11 fusion approaches were fine-tuned from other trained CNNs for trip hazard detection.

\subsection{Visual Classifer Overview} \label{section:expsetup}

A Convolutional Neural Network (CNN) performing semantic segmentation is used as a visual classifier to detect trip hazards. Here the non fusion trip detecting networks (RGB, Depth and HHA) use the VGG architecture \cite{Simonyan2014a} adopted into a Fully Convolutional Neural Network (FCN) for semantic segmentation (or dense classification) as per Long et. al \cite{Long2014}. However, for the fusion approaches, various modifications to the VGG architecture are required, which will now be explained.

\begin{figure}[t!]
	\includegraphics[width=0.91\linewidth]{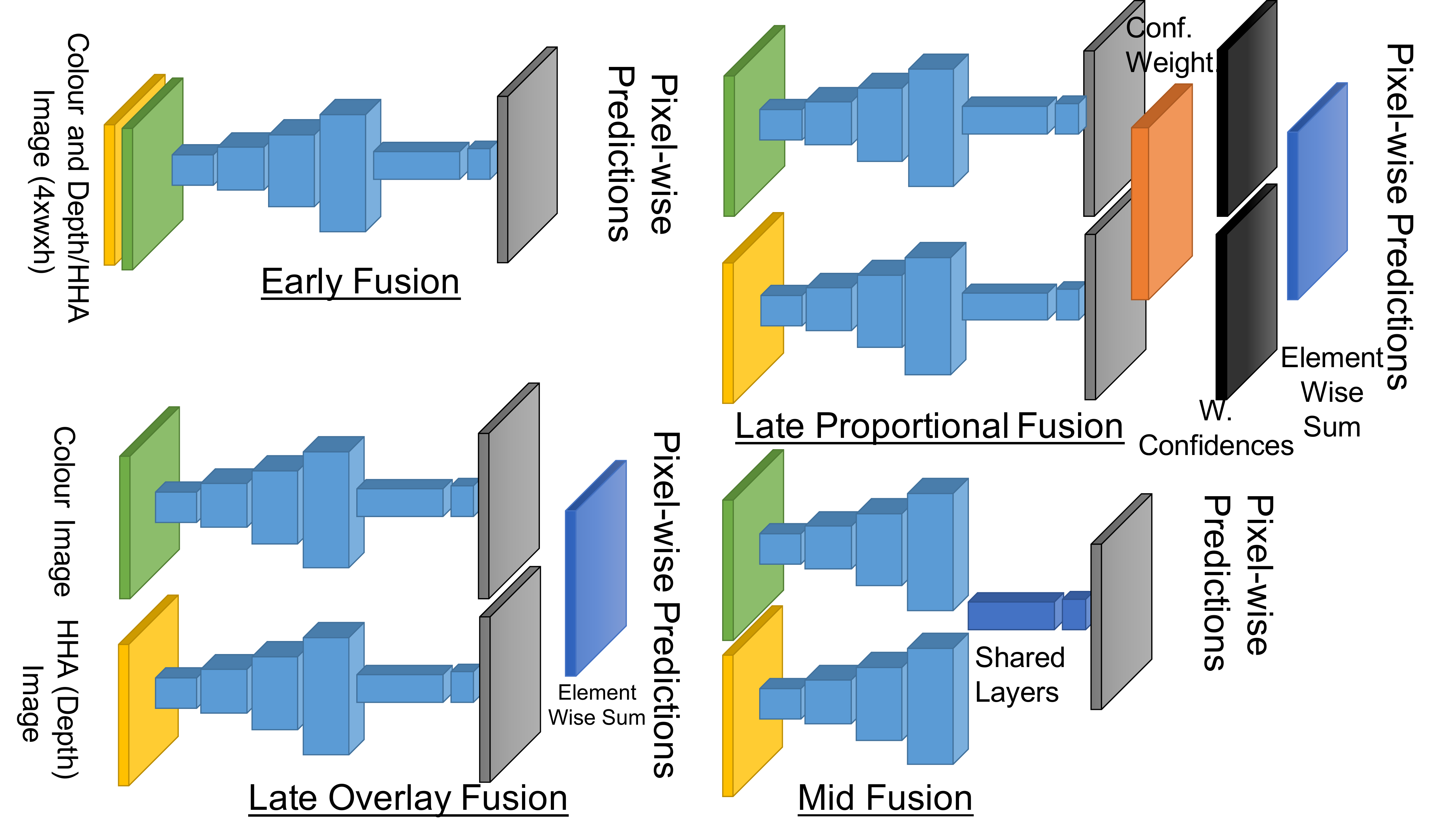}
	\centering
	\caption{Early (top left), Late Overlay (bottom right), Late Proportional (top right), and the Mid Level Fusion approaches (bottom right). Late Overlay Fusion naively combines the final predictions of the colour and depth/HHA modalities. Late Proportional Fusion weights predictions based on their relative confidence which are then summed together \cite{Ge2016}. Mid Fusion combines the activations from each modality's final pooling layer, followed by two convolutional layers and then upscaled trip hazard predictions.}
	\label{fig:all_fusion}
	\vspace{-0.5cm}
\end{figure}

\subsection{Fusion Architectures} \label{section:classiferArch}
This research investigates four different colour and depth modality fusion approaches; Early, Late Overlay, Mid and Late Proportional Fusions. Each of these fusion approaches will be trained on two image combinations; colour with raw depth and colour with HHA images.

First, the Early Fusion approach will concatenate colour with depth and colour with HHA images, to be passes as inputs to the network; see Figure~\ref{fig:all_fusion} (top left).

The second fusion approach, Late Overlay Fusion, will overlay trip detections from two single modality networks by naively adding the two final classifications together where depth is recorded, otherwise the colour networks prediction will be used; see Figure~\ref{fig:all_fusion} (bottom left). The two single modality networks will be trained separately, then used for testing Late Overlay fusion.

The third fusion approach, Mid Fusion, seen in Figure~\ref{fig:all_fusion} (bottom right), is to combine the two modalities by concatenating the activations from each final pooling layer. These concatenated activations will then be passed through the final two convolutional layers before upscaling for the final pixel-wise trip hazard predictions.

Finally the fourth fusion approach, Late Proportional Fusion, visualised in Figure~\ref{fig:all_fusion} (top right), is based on a network ensemble approach called MixDCNN \cite{Ge2016}. This research introduces, and then investigates, this approach for modality fusion of colour with depth and HHA images.
Late Proportional Fusion weights the two predictions by the relative performance. More details about the MixDCNN approach can be found in the original text \cite{Ge2016} and code (http://tinyurl.com/jwtjk35).

\section{Construction Site Dataset} \label{section:data}
This section establishes how the colour and depth image data of the construction site was collected, then labelled, for supervised training and testing of the different modality fusion approaches. Subsection~\ref{section:DataCollection} describes the techniques used for data collection followed by subsection~\ref{section:DataLabelingPreprocessing} which describes data labelling.

\begin{figure}
	\centering
	\setlength\tabcolsep{1.0pt}
	\begin{tabular}{cccc}
		\centering
		\subfloat[Gnd Floor]{\includegraphics[width=0.23\linewidth]{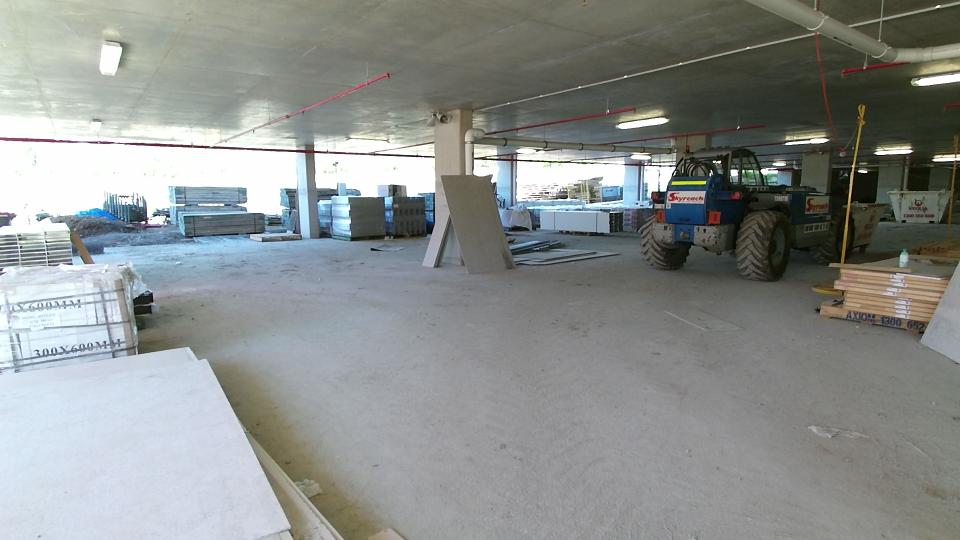}} &
		\subfloat[2nd Floor]{\includegraphics[width=0.23\linewidth]{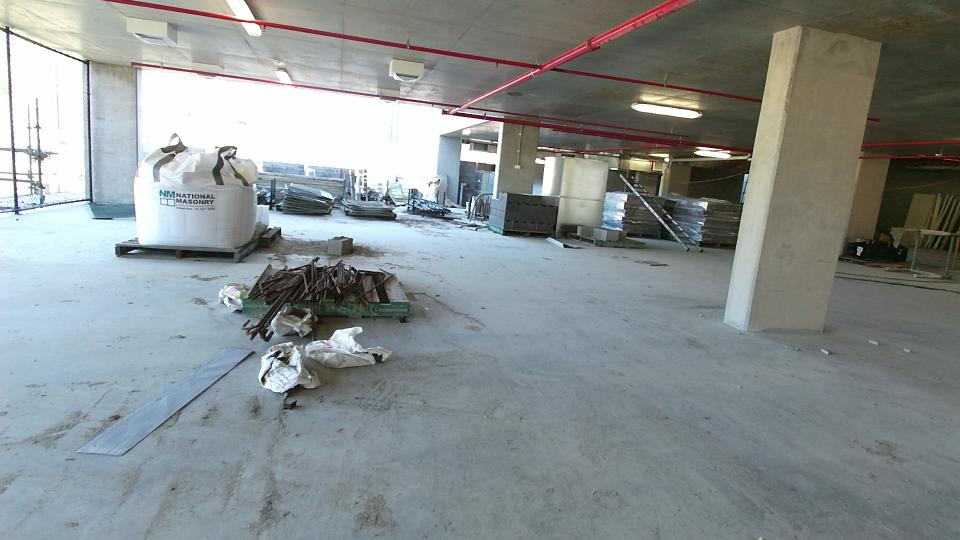}} &
		\subfloat[4th Floor]{\includegraphics[width=0.23\linewidth]{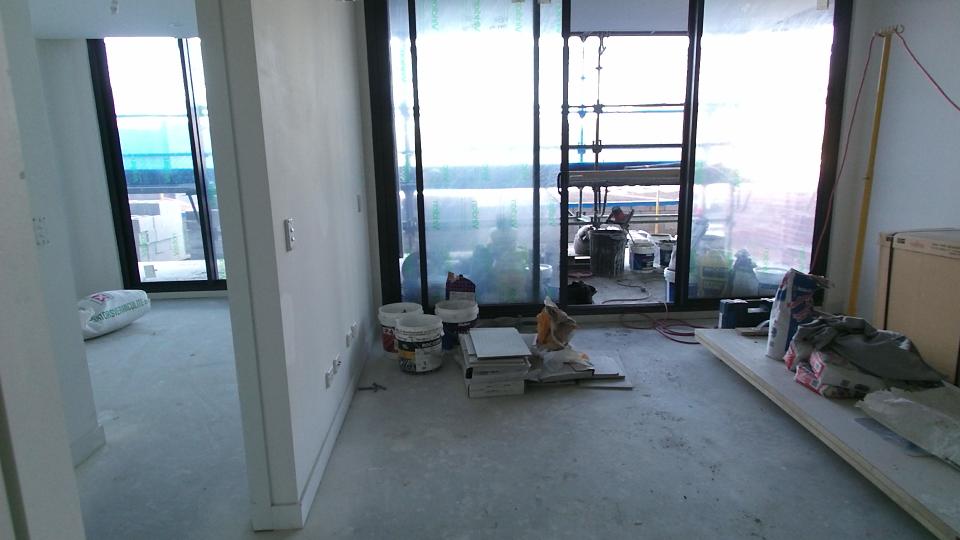}} &
		\subfloat[7th Floor]{\includegraphics[width=0.23\linewidth]{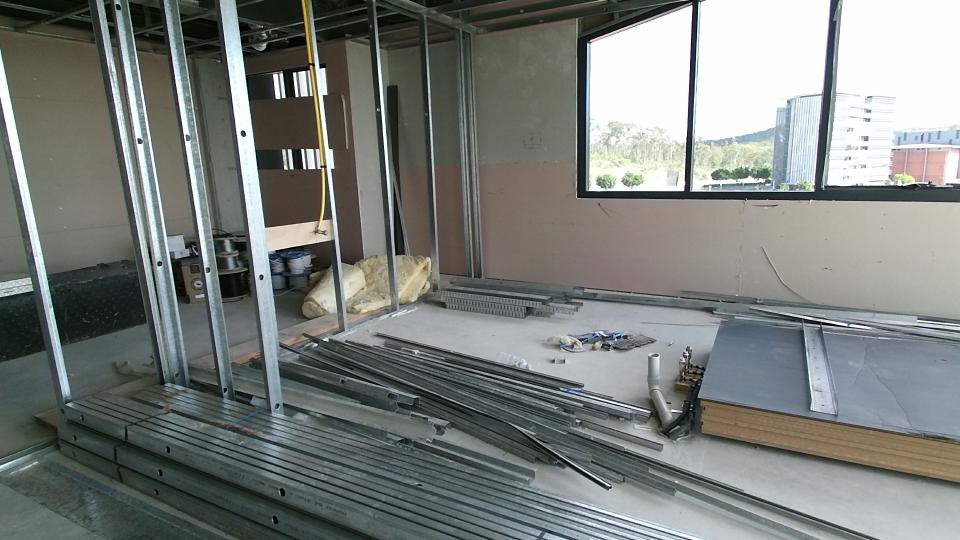}}
	\end{tabular}
	\caption{Sample colour images from each of the four floors recorded in our dataset.}
	\label{fig:datasetSample}
	\vspace{-0.5cm}
\end{figure}

\begin{figure}
	\centering
	\setlength\tabcolsep{1.0pt}
	\begin{tabular}{ccc}
		\centering
		\subfloat[Depth Image \newline(colourised)]{\includegraphics[width=0.23\linewidth]{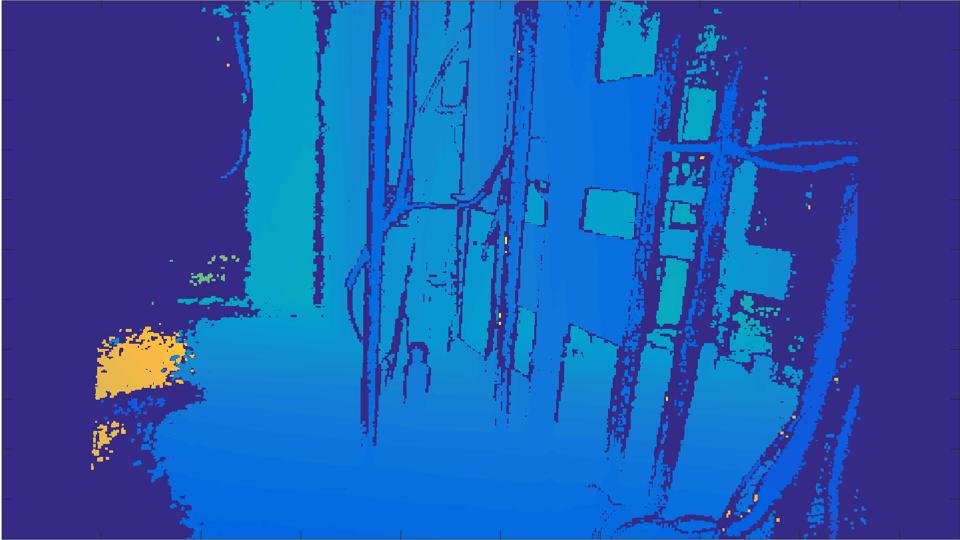}} &
		\subfloat[Colour Image]{\includegraphics[width=0.23\linewidth]{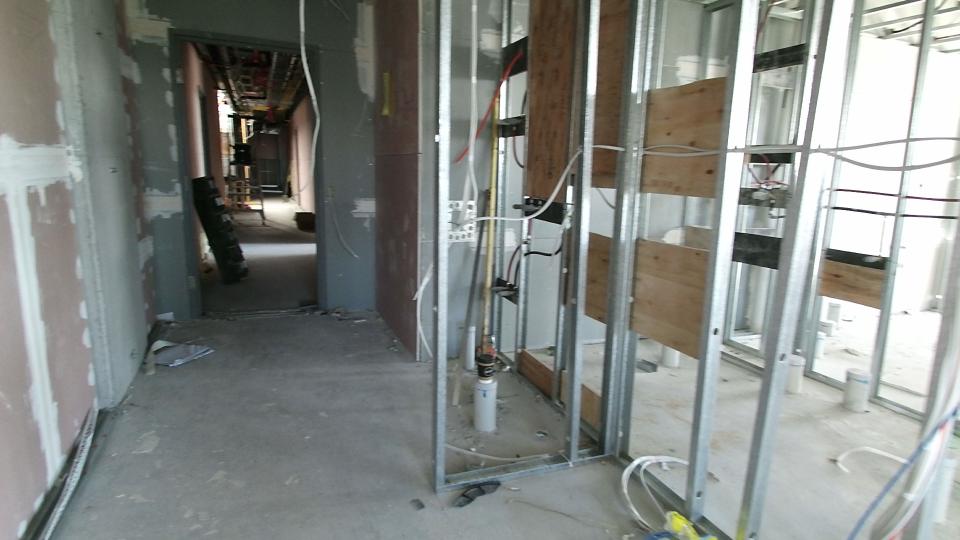}} &
		{\subfloat[HHA Image]{\includegraphics[width=0.23\linewidth]{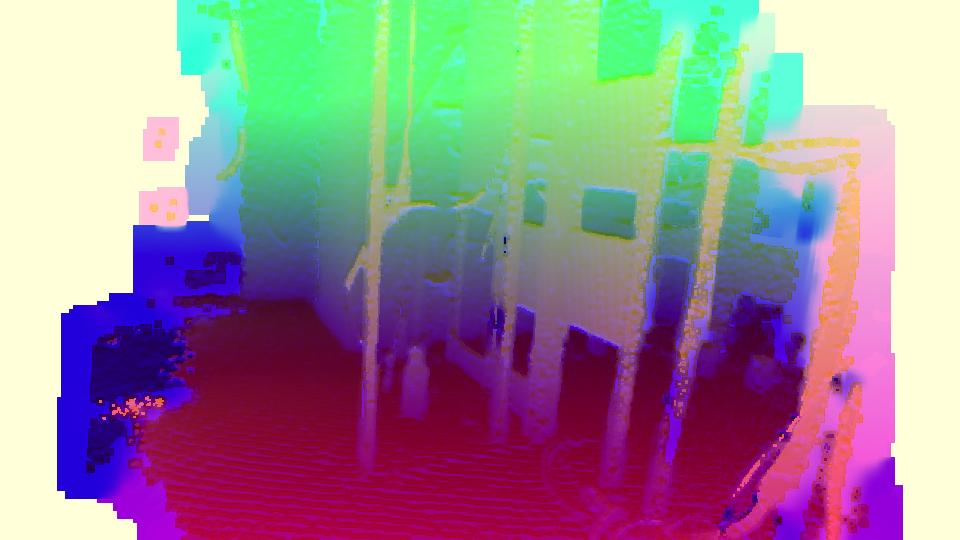}}}
	\end{tabular}
	\caption{(a) and (b) {presents the} two types of visual data recorded. (a) is a sample raw depth image colourised for visualisation purposes, and (b) is the corresponding colour image. (c) presents a sample HHA image which is generated from the raw depth images.}
	\label{fig:geometricdatasetSample}
	\vspace{-0.5cm}
\end{figure}
\begin{figure}
\vspace{-0.5cm}
	\centering
	\includegraphics[width=0.7\linewidth]{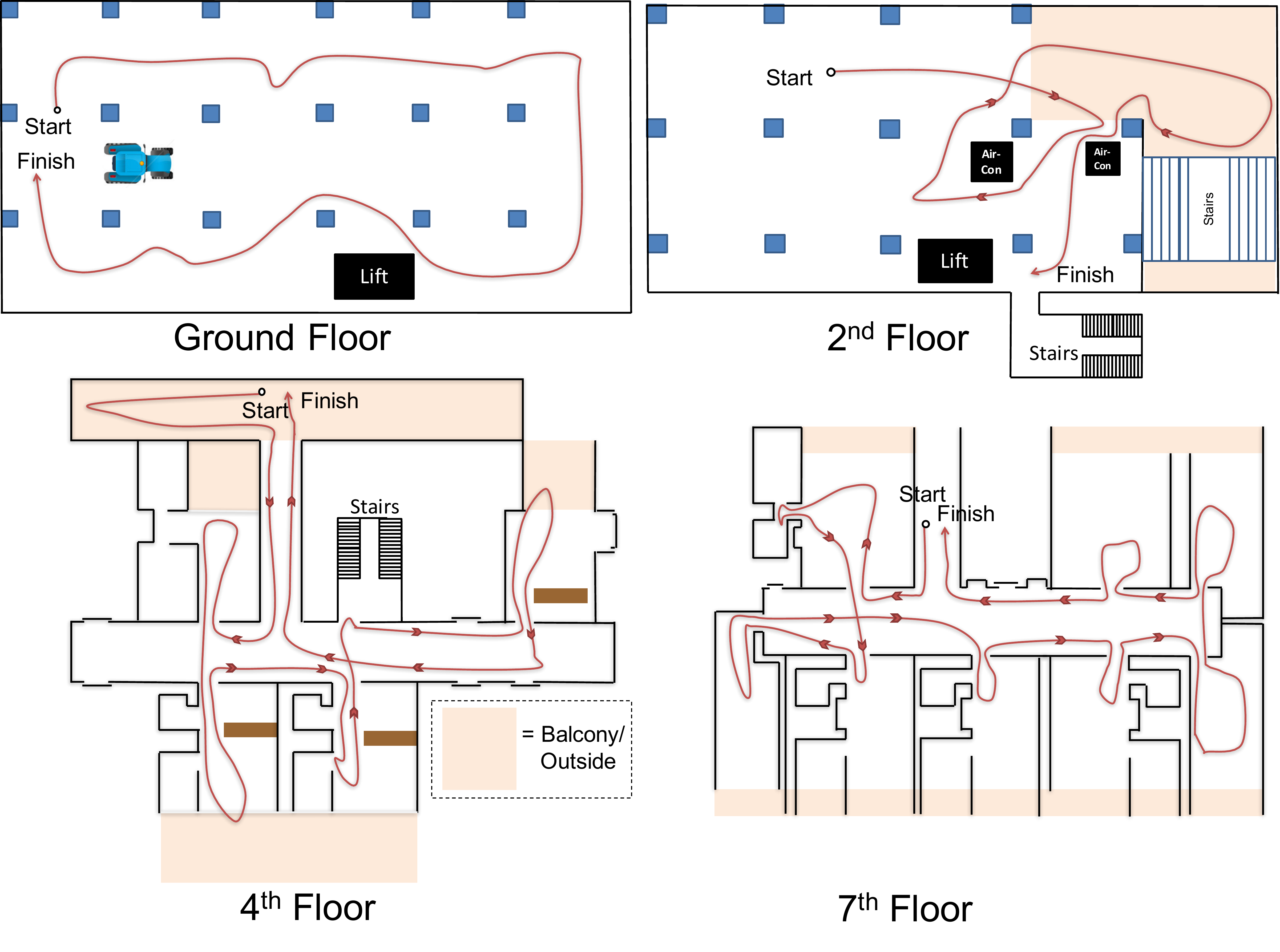}
	\caption{Plans of each floor of the construction site recorded. The red line is the traversal made when recording with the Kinect 2 RGB-D camera, the blue squares are support pillars.}
	\label{fig:gndFloorPlan}
\end{figure}

\subsection{Dataset Collection} \label{section:DataCollection}

This research introduces a visual dataset of labelled colour, raw depth and HHA images of a construction site. The construction site recorded is an apartment complex four months from completion, with nine floors in total. Four of these floors were recorded in the visual dataset covering approximately 500m\textsuperscript{2} each. When {completed, two} of the floors will be a car park and the other two floors will contain apartments. Example images can be seen in Figure~\ref{fig:datasetSample} and the floor plans of each floor with the traversals made can be seen in Figure~\ref{fig:gndFloorPlan}. The data was collected to train, and then test the trip hazard detector. It has also been made public to encourage further research in this area (\textit{http://tinyurl.com/Tripdataset}).

Our visual dataset improves on an earlier dataset \cite{mcmah15} in two key aspects. First it covers a larger area; the dataset from earlier work covered approximately 300m\textsuperscript{2} of one site and 100m\textsuperscript{2} in another, while our new dataset covers approximately 2000m\textsuperscript{2} across the four floors recorded. The second aspect is the use of a depth sensor to obtain depth images. Earlier work only contained colour images using a GoPro camera \cite{mcmah15}, however this research uses a Microsoft Kinect 2 camera which can record both colour and raw depth images.

To collect this new dataset, the Kinect 2 was mounted to a tripod with a battery pack. The Kinect 2 was set to record video at a resolution of 540 x 960 for both depth and colour images due to computational limitations. The video was recorded while traversing a loop of each floor stopping approximately every 10 meters to do a short pan of the area.

\subsection{Dataset Labelling} \label{section:DataLabelingPreprocessing}

With the data recorded, the colour and depth videos were split into individual images and then labelled (ground truthed). Images were extracted every 0.5 seconds of video recording. In order for the human to label the extracted images, a trip hazard was defined as an entire or significant part of an object less than 0.5 meters above the ground plane which would cause a human to trip or stumble. Using this definition, every instance of a trip hazard recorded along the traversals in Figure~\ref{fig:gndFloorPlan} was labelled. If the view point on a trip hazard in an image changed (i.e. approaching the hazard from the opposite side) it was labelled again. The software used to label trip hazards in the images was a MATLAB script called the Object Labelling Tool, by Hoiem \cite{Hoiem}. This MATLAB script allows users to draw polygons over regions of interest, which in this case were regions of the images containing instances of trip hazards.

\section{Network Training Procedure} \label{section:networkTraining}
This section describes how the fusion and non fusion FCNs will be trained for trip hazard detection using a type of training called fine-tuning. { Fine-tuning is where a network has already been trained (pre-trained) to perform a similar, but different task. The training methodology is split into four areas;  HHA Preprocessing, Weight Initialisation, Fine-tuning, Hyperparemeter Search and Cross-Validation.}

\subsubsection{HHA Preprocessing}
To help the fusion and non fusion networks better learn features from depth information, the raw depth images were preprocessed to generate HHA images. HHA images are a three channel depth image, where each channel of this image contains the Horizontal disparity, Height above ground, and the Angle the pixel normal makes with an inferred gravity direction. Gupta et al. \cite{Gupta2014a} proposed this encoding, and found it gave improved detection results with CNNs. The HHA images were generated with the functions from the RCNN-Depth toolbox provided by \cite{Gupta2014a}. However, a modification was made to the height above ground channel because the assumption made by \cite{Gupta2014a}, where the ground plane is the lowest point in the image, was found to be invalid for our depth data. This was due to noise from the K2 causing some points to be too low to be valid. To reduce this error, an additional constraint was added which limited the ground plane to a maximum of 1.9 meters below the camera. These values were chosen because the camera was held 1.8 meters above the ground and an extra 10cm was added to the limit to account for any additional vertical displacement of the camera during recording.

\subsubsection{Weight Initialisations}
The weights of all the fusion and non-fusion FCNs were initialised from networks pre-trained per Long et al. (2015) \cite{Long2014} on the NYU dataset \cite{Silberman2012}.
All colour weights were initialised from a network pre-trained on NYU colour images, and all the depth and HHA networks were initialised from a network pre-trained on NYU with HHA images. For the early fusion approach, the weights of the first layer were initialised from both the colour and HHA pre-trained networks, and then the rest of the network was initialised from the colour network. The shared layers of the Mid Fusion were trained from scratch. All networks were initialised from the same two networks to keep comparisons between approaches impartial, and prevent the fusion networks from relying on a single modality.

\subsubsection{Fine-tuning}
{With the weight initialisation methods chosen, the FCNs are then fine-tuned by replacing the final convolution layer with a new layer for trip hazard detection. The Cross Entropy Loss Function was used with a Softmax Classifier to fine-tune all approaches investigated. Because FCNs output multiple predictions per image, the per-pixel losses were added together and then used for Stochastic Gradient Descent (SGD) optimisation during training with a batch size of 1. }

\subsubsection{Hyperparameter Search}
{To facilitate more effective results from fine-tuning we performed a grid search of key hyperparameters. Due to time and computation constraints, a selection of parameters were searched over a specific range. Two fixed learning rates of $1^{-10}$ and $1^{-12}$, as well as two learning rate multipliers for the final convolutional layer (5 and 10) were experimented with for all approaches. For Early Fusion, a search was done over two learning rate multipliers for the first convolutional layer; 4 and 10, and a fixed final layer multiplier of 5. For Mid Fusion an additional search was conducted over the learning rate multipliers for the shared layers (2 and 5) and the Dropout ratios (0.5 and 0.75). During the searches we kept momentum at 0.99 and doubled all learning rates for the biases. The hyperparameters giving the lowest loss on the validation set were used for testing with cross-validation, displayed in Table}~\ref{table:overallresults}.

\subsubsection{Cross-Validation}
With the networks initialised and a hyperparameter search completed, the FCNs are then trained using cross-validation of the four floors of the construction site. To do this we partitioned the data into 4 folds, testing on each floor while training on the other three. The performance metrics from testing on each fold were then averaged together for the results shown in Section~\ref{section:eval}.

\section{Evaluation} \label{section:eval}
The results of the FCNs trained to detect trip hazards on the new construction site dataset are displayed in Figures~\ref{fig:PR_curve_nonfusion}, ~\ref{fig:PR_curve_fusion} and~\ref{fig:Trip_IOU_curve_fusion} as well as Tables~\ref{table:overallresults} and~\ref{table:tripnet_comp}. The text below first establishes the performance metrics used, followed by an analysis of the results.

\subsection{Performance Metrics}
Five common performance metrics were used to evaluate the trip detectors; Precision, Recall, F1-score, Trip IOU and Trip Object Detection. {Precision measures the number of correct trip predictions relative to the total number of true and false trip predictions made, with higher values being better. It is defined as }$\frac{T_{p}}{T_{p} + F_{p}}$, with $T_{p}$ as the True Positives, $F_{p}$ the False Positives and $F_{n}$ the False Negatives. Recall measures the percentage of true trip predictions over the total number of trip hazards, with a higher recall better (defined as $\frac{T_{p}}{T_{p} + F_{n}}$). F1-Score is the harmonic mean of precision and recall, providing a summary of both with a higher value better (defined as $2\cdot\frac{P \cdot R}{P+R}$). Trip Intersection Over Union (IOU) measures how well the image regions predicted to be trips overlap with the ground truth regions, higher is better (defined as $\frac{ T_{p} }{ T_{p} + F_{n} + F_{p} }$). The Trip Object Detection metric is the number of correctly predicted trip hazard objects, divided by the total number of ground truth trip objects. This metric could be used to deploy this system on construction sites, and inform workers of regions where trip hazards have been detected.

\subsection{Results And Analysis}

From the raw results, our investigation reveals the following key insights:
\begin{itemize}
    \item Multi-modal trip hazard detection that fuses RGB and HHA image information is superior to a single modality system.
    \item The Late Proportional Fusion approach proved to be superior compared to all other tested fusion approaches.
    \item Exploiting structural information via the HHA encoding is superior to using raw depth images.
 	\item Appearance information via colour images is superior for trip detection to structural information via HHA and raw depth images
 	\item Improved performance reported in our updated approach using colour over an earlier prototype TripNet \cite{mcmah15}, also using colour.
\end{itemize}

\subsubsection{Multi-modal Hazard Detection Outperforms Single-Modality Detection}

The investigation shows quantifiable improvements over single modality approaches by fusing appearance based colour information with structural depth information. As illustrated in the precision-recall curve in Figure~\ref{fig:PR_curve_nonfusion} and Table~\ref{table:overallresults}, the best fusion approach (Late Proportional Fusion of RGB and HHA images) outperforms the three single modality approaches (RGB, HHA and Depth).

The increase in performance can be explained by the complementary nature of colour and depth (HHA) information. While a large number of trip hazards are detected based on colour information alone, there are cases where depth information can help resolve ambiguity and increase detection precision. This can be seen in Figures~\ref{fig:late_prop_comparison_156} and~\ref{fig:late_prop_comparison_74} where our Late Proportional Fusion approach leverages information from RGB and HHA images to better detect trip hazards. In these cases the HHA based detector outperforms the RGB based detector in instances with large nearby trip hazards.

However the performance boosts from modality fusion are not as significant as some might expect (Figures~\ref{fig:PR_curve_fusion} and~\ref{fig:Trip_IOU_curve_fusion}). One explanation for this are instances where the RGB and depth/HHA predictions overlap, as seen in Figure~\ref{fig:late_prop_depth_no_effect_99}. Another are instances where depth/HHA based detections fail to see certain hazards, illustrated in Figure~\ref{fig:rgb_vs_hha_res_range}.

\newcommand{\CurveLM}{0.95}
\begin{figure}[t]
	\centering
	\includegraphics[width=\CurveLM\linewidth]{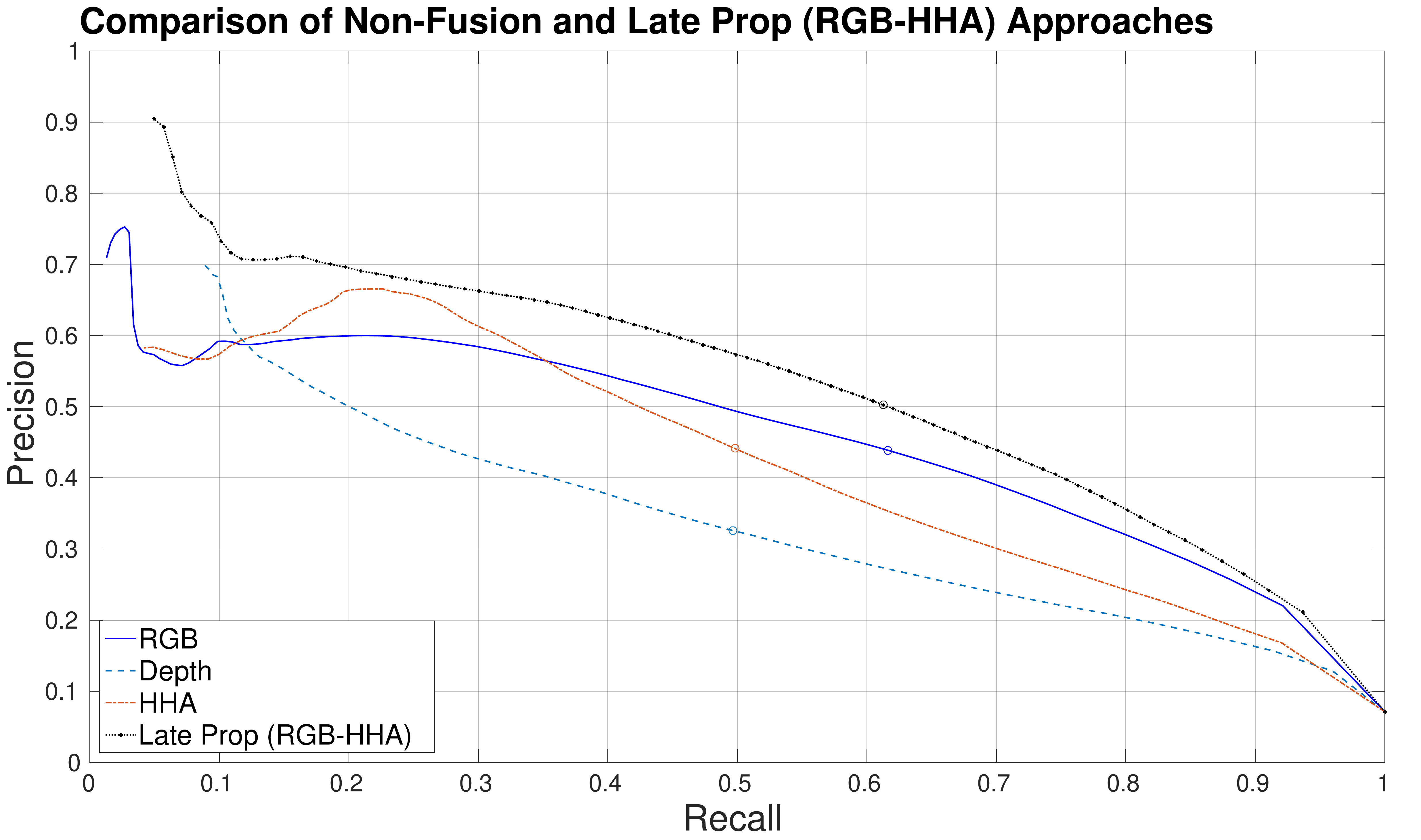}
	\caption{Precision Recall Curve comparing the non fusion approaches, circles show operating points displayed in Table~\ref{table:overallresults}. Curves closer to the top right show higher performance, as they achieve higher precision and recall.}
	\label{fig:PR_curve_nonfusion}
	\vspace{-0.2cm}
\end{figure}

\begin{figure}[t]
	\centering
	\includegraphics[width=\CurveLM\linewidth]{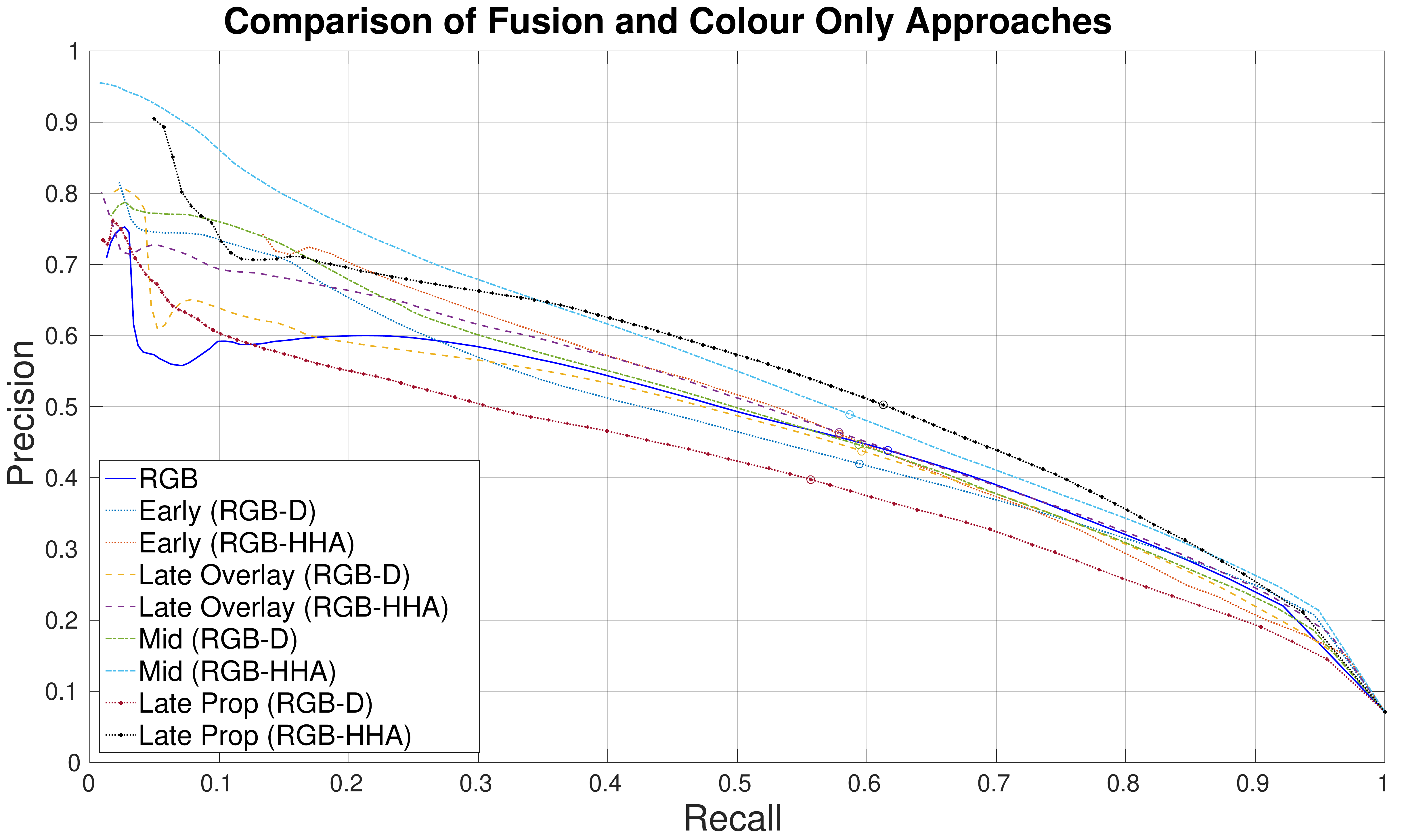}
	\caption{Precision Recall curve comparing the fusion approaches; circles show the operating points in Table~\ref{table:overallresults}.}
	\label{fig:PR_curve_fusion}
	\vspace{-0.3cm}
\end{figure}

\begin{figure}[t]
	\centering
	\includegraphics[width=\CurveLM\linewidth]{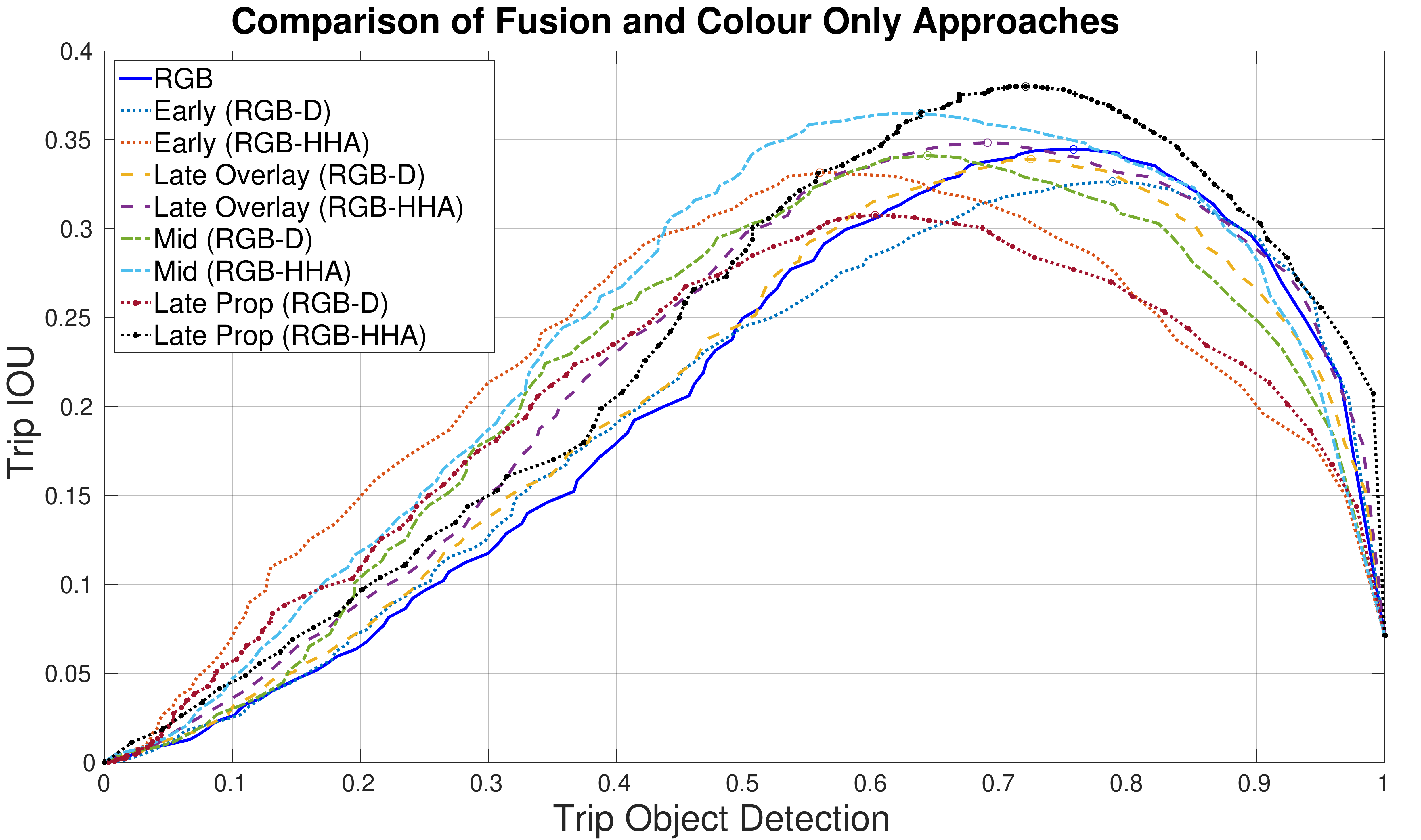}
	\caption{Curve showing the trade off between Trip Object Detection and Trip IOU. Circles show the operating points seen in Table~\ref{table:overallresults}. Curves closer to the top right show higher performance.}
	\label{fig:Trip_IOU_curve_fusion}
	\vspace{-0.5cm}
\end{figure}

\begin{table}[!b]
\vspace{-0.3cm}
\centering
\caption{The 11 different fusion and non fusion approaches. Results here are the averaged performance from cross validation. The metrics shown are from the operating points which achieve the highest F1-score for each approach.}
\begin{tabular}{@{}p{0.8cm}cccccc@{}}
\toprule
             \multirow{2}{*}{Fusion}  & \multirow{2}{*}{Modality}    & \multirow{2}{*}{Prec}& \multirow{2}{*}{Rec}&\multirow{2}{*}{F1}& Trip         & Trip Obj \\
                 &        &                      &                     &                   & IOU          & Det. \\
\midrule
\multirow{3}{0.8cm}{None}& Depth                   & 32.6                 & 49.6                & 39.2              & 24.7         & 58.1  \\ 
& HHA                      & 44.1                 & 49.8                & 46.2              & 30.5         & 50.8  \\ 
& RGB                       & 43.9                 & \textbf{61.6}       & 50.8              & 34.5         & 75.7  \\
\midrule
\multirow{2}{0.8cm}{Early} & RGB-D            & 42.0                 & 59.4                & 48.5              & 32.6         & \textbf{78.7} \\ 
 & RGB-HHA          & 46.2                 & 57.9                & 49.8              & 33.2         & 55.9 \\ 
\midrule
\multirow{2}{0.8cm}{Late Overl.} & RGB-D & 43.8                 & 59.6                & 50.3              & 33.9         & 72.3  \\ 
& RGB-HHA & 46.4                 & 57.9                & 51.2              & 34.8         & 70.0 \\
\midrule
\multirow{2}{0.8cm}{Mid} & RGB-D     & 44.6                 & 59.4                & 50.7              & 41.1         & 64.3\\ 
 & RGB-HHA     & 48.9                 & 58.7                & 53.1              & 36.5         & 63.8\\
\midrule
\multirow{2}{0.8cm}{Late Prop.} & RGB-D      & 39.8                 & 55.6                & 46.2              & 30.8         & 60.2 \\
 & RGB-HHA    & \textbf{50.2}        & \textbf{61.3}       & \textbf{54.8}     & \textbf{38.0}& 72.0 \\
\bottomrule
\end{tabular}
\label{table:overallresults}
\end{table}

\begin{table}[!b]
\vspace{-0.3cm}
	\centering
	\caption{Comparison of proposed colour (RGB) and HHA trip hazard detectors with the earlier prototype detector, TripNet \cite{mcmah15}. Results in the Table were obtained by testing the networks on the ground floor of the construction site and training on the other three; without cross-validation.}
	\begin{tabular}{@{}lccccc@{}}
		\toprule
		& \multirow{2}{*}{Prec}& \multirow{2}{*}{Rec}&\multirow{2}{*}{F1} & Trip         & Trip Obj \\
		&                      &                     &                    & IOU          & Detection \\
		\midrule
		TripNet \cite{mcmah15}   & 17.2                 & \textbf{76.1}       & 19.8               & 16.3         & 67.7 \\

		HHA                      & 56.4                 & 50.9                & 53.5               & 36.5         & 50.3  \\ 

		RGB                      & \textbf{56.7}         & 57.0                & \textbf{56.9}     & \textbf{39.7}& \textbf{71.4}  \\		\bottomrule
	\end{tabular}
	\label{table:tripnet_comp}
\end{table}

\newcommand{\VLM}{0.46}

\begin{figure}
    \centering
\begin{tabular}{cc}
    \subfloat[ ]{\includegraphics[width=\VLM\linewidth]{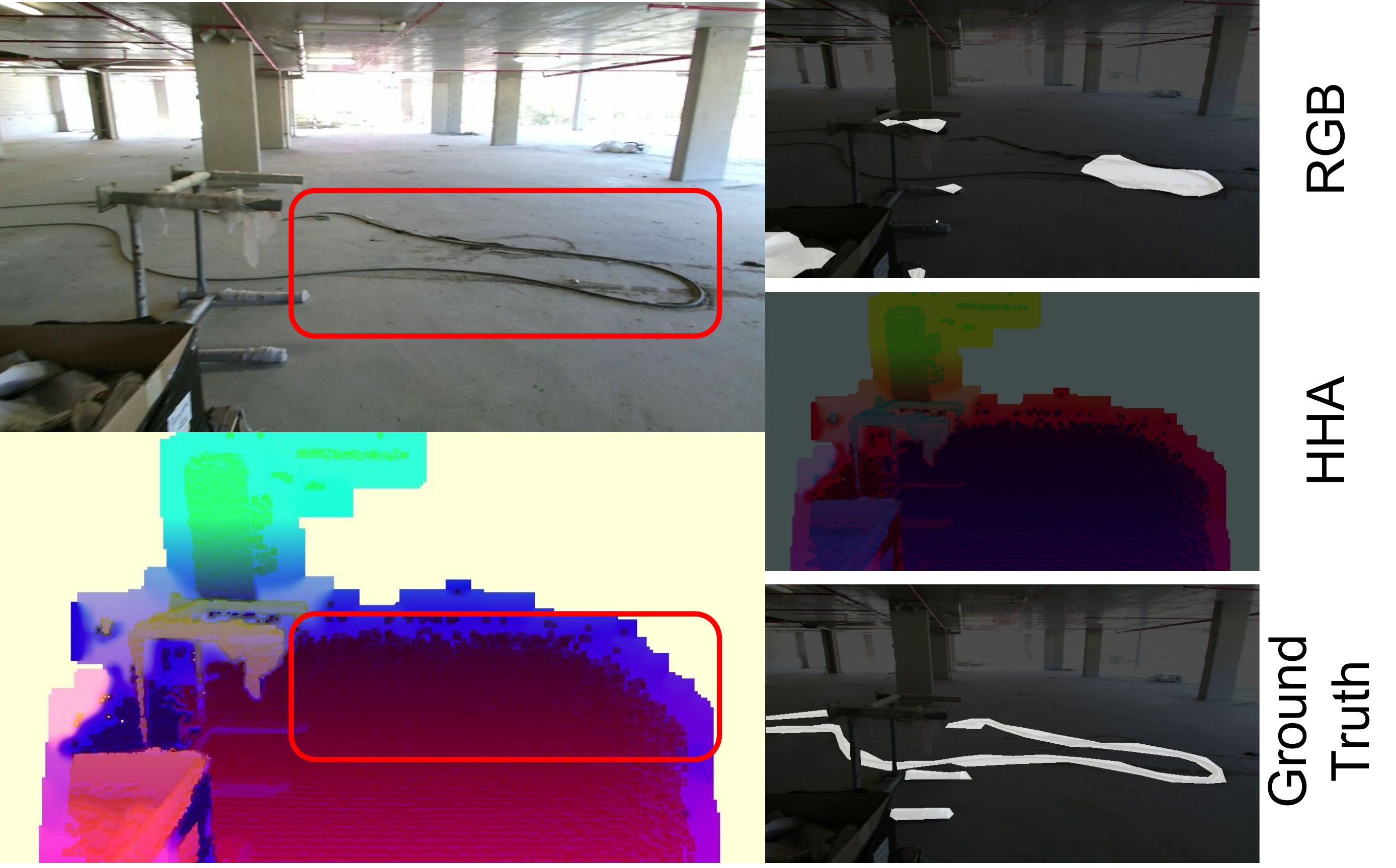}} &
    \subfloat[ ]{\includegraphics[width=\VLM\linewidth]{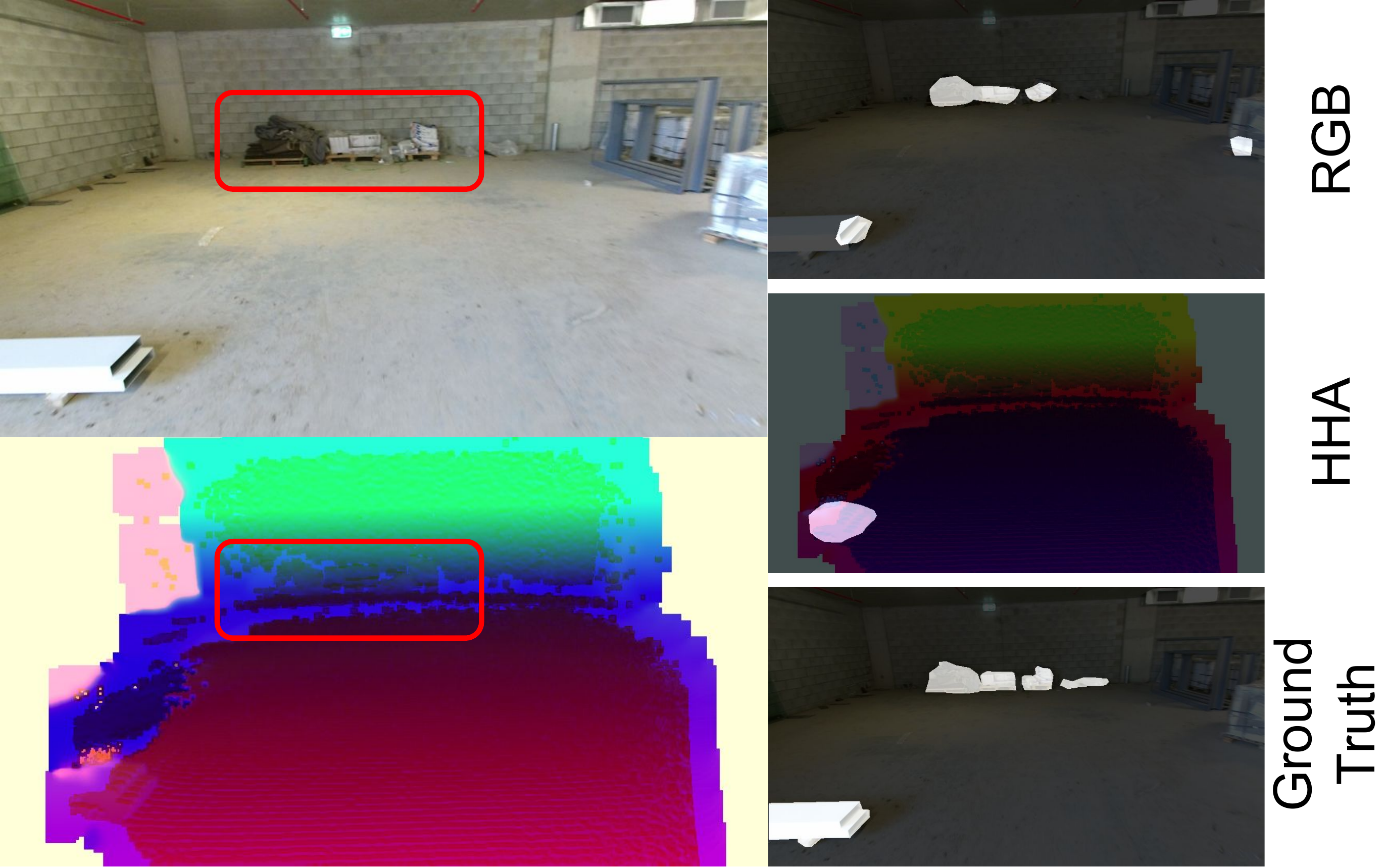}}
\end{tabular}
\caption{{(a) The resolution of the depth sensor makes it more difficult for the trip detector to see smaller trip hazards, such as cables. (b) Limited range of depth information restricts the trip detection capability of the HHA trained FCN compared to the RGB. The right hand columns of (a) and (b) shows trip segmentations from the RGB and HHA FCNs as well as the ground truth labels.}}
\label{fig:rgb_vs_hha_res_range}
\vspace{-0.5cm}
\end{figure}

\newcommand{\LM}{0.18}
\begin{figure*}
	\centering
	\setlength\tabcolsep{1.0pt}
\begin{tabular}{ccccc}
	\subfloat[Colour Image]{\includegraphics[width=\LM\linewidth,clip]{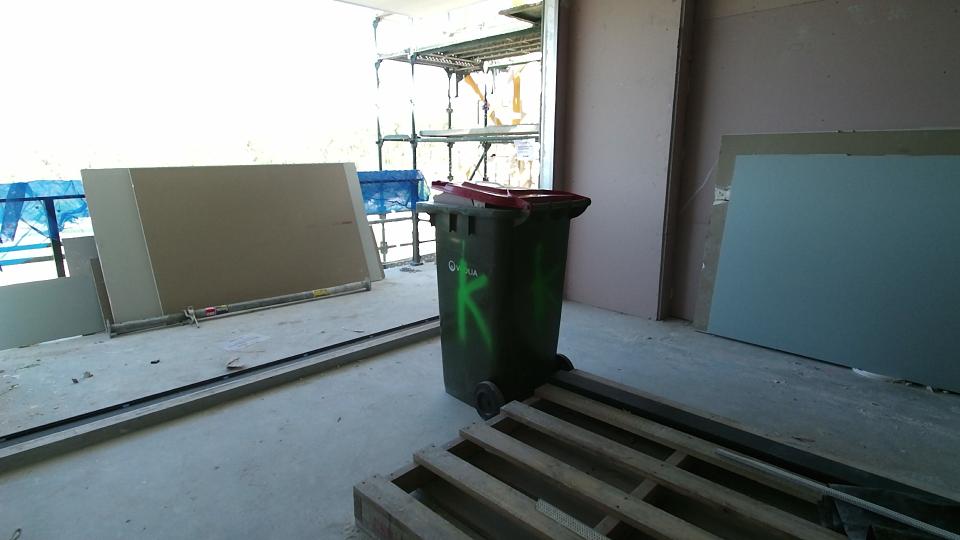}} &
	\subfloat[Ground Truth]{\includegraphics[width=\LM\linewidth,clip]{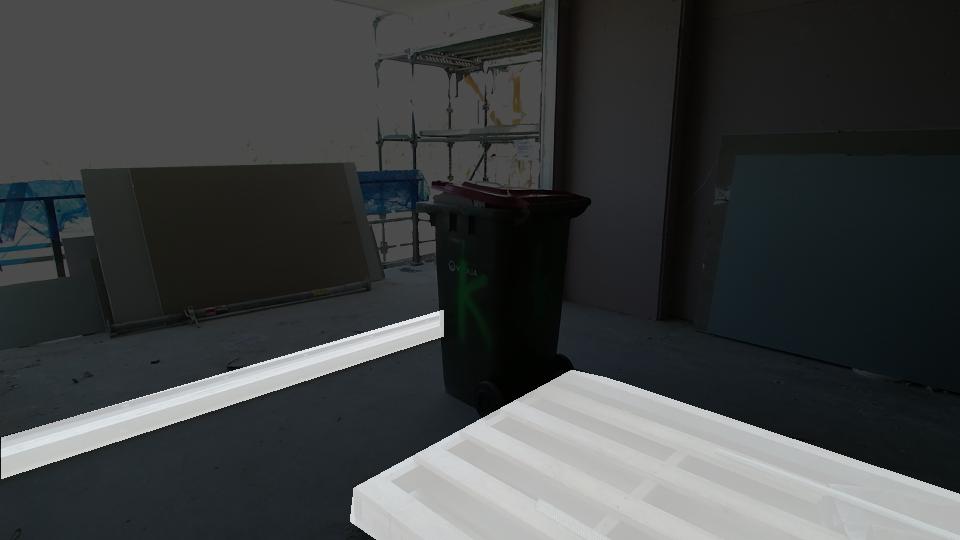}} &
	\subfloat[RGB]{\includegraphics[width=\LM\linewidth,clip]{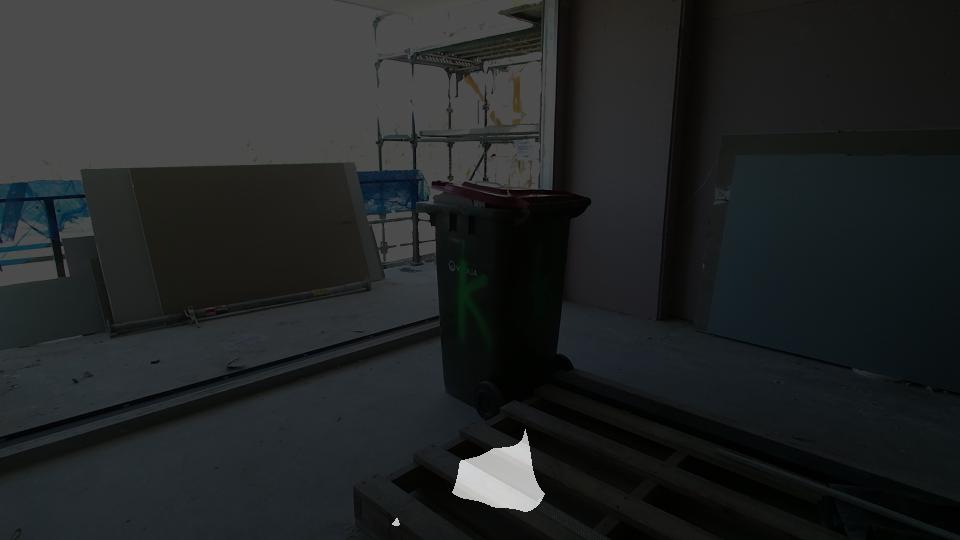}} &
    \subfloat[HHA]{\includegraphics[width=\LM\linewidth,clip]{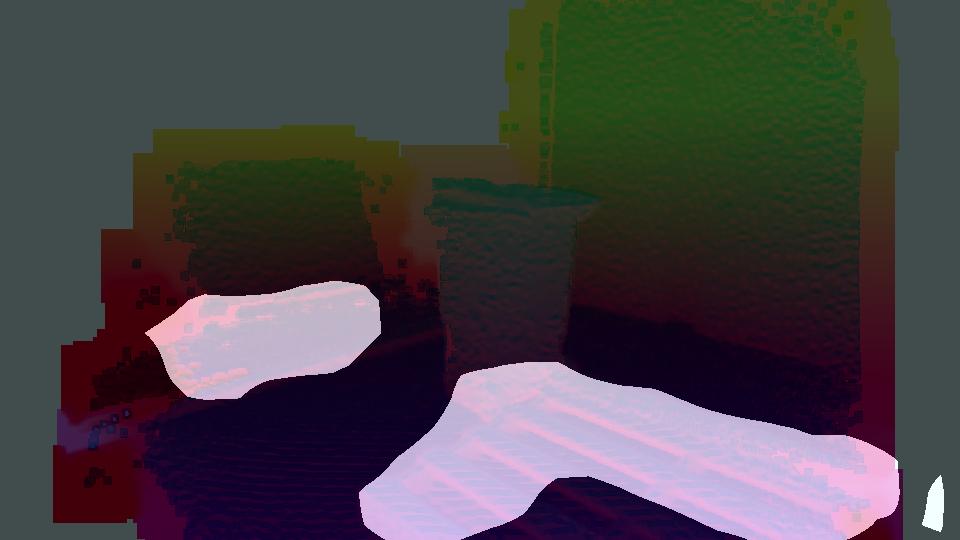}} &
    \subfloat[Late Prop RGB-HHA]{\includegraphics[width=\LM\linewidth,clip]{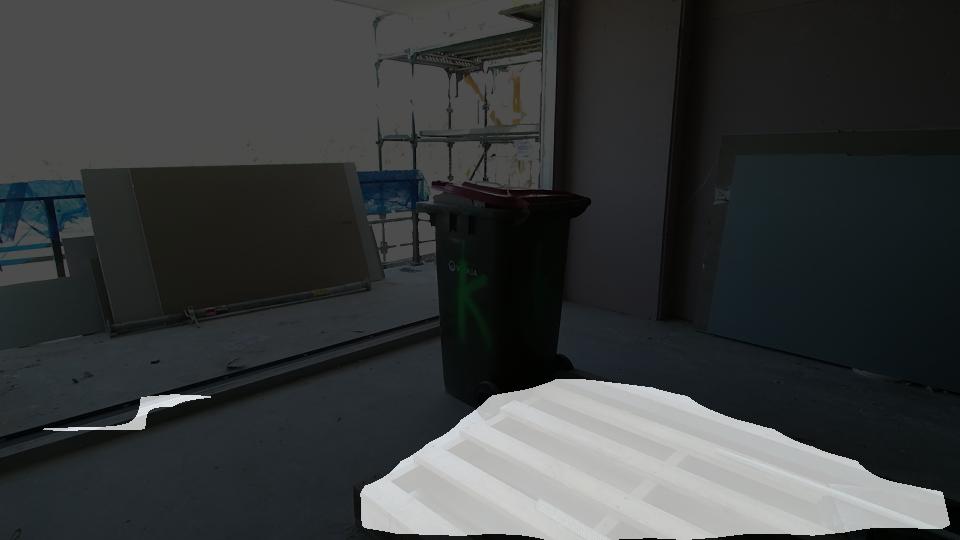}}
\end{tabular}
\caption{Improvements from Late Proportional Fusion (e). Here a large nearby trip hazard, the pallet in the bottom right corner, is within the sensing range of the depth sensor and therefore capable of being detected by the HHA trained trip detector. }
\label{fig:late_prop_comparison_156}
\vspace{-0.5cm}
\end{figure*}

\begin{figure*}
	\centering
	\setlength\tabcolsep{1.0pt}
	\begin{tabular}{ccccc}
		\subfloat[Colour Image]{\includegraphics[width=\LM\linewidth,clip]{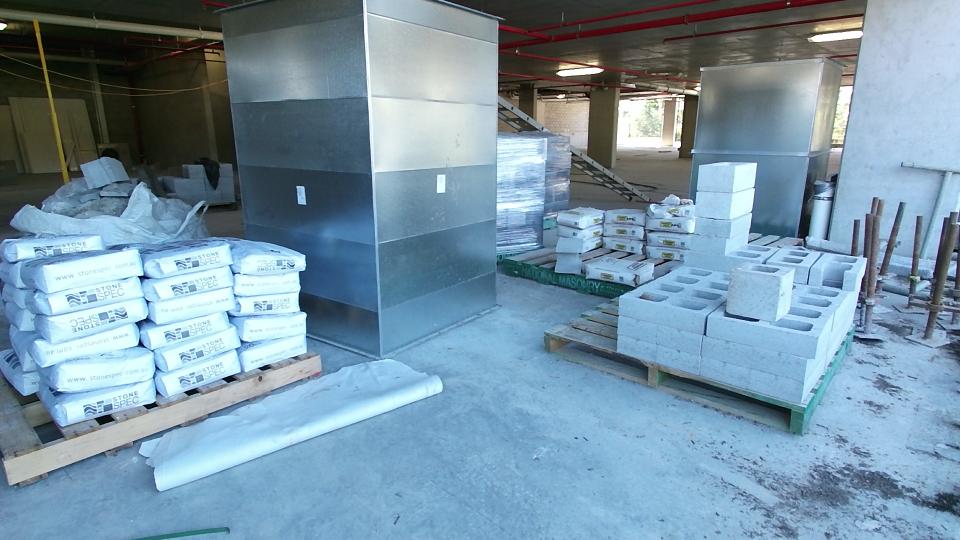}} &
		\subfloat[Ground Truth]{\includegraphics[width=\LM\linewidth,clip]{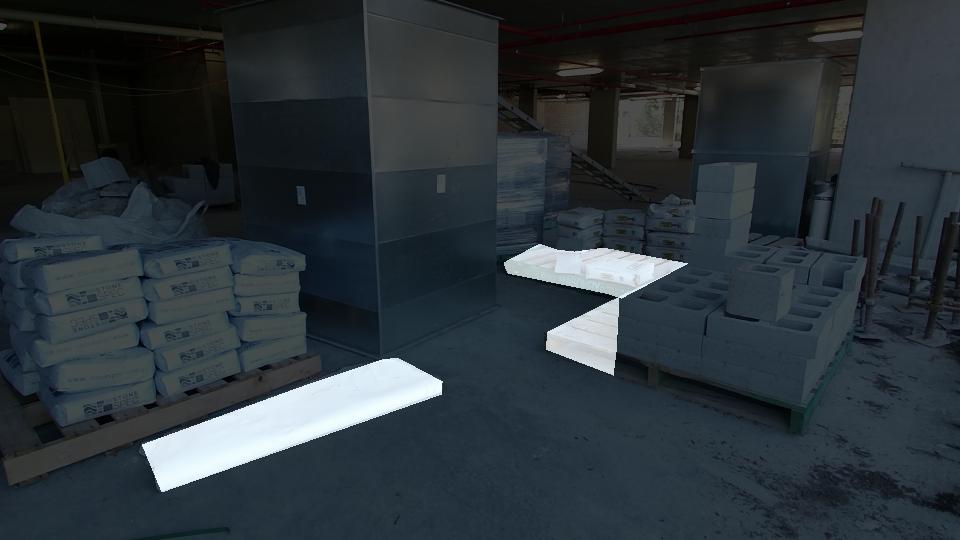}} &
		\subfloat[RGB]{\includegraphics[width=\LM\linewidth,clip]{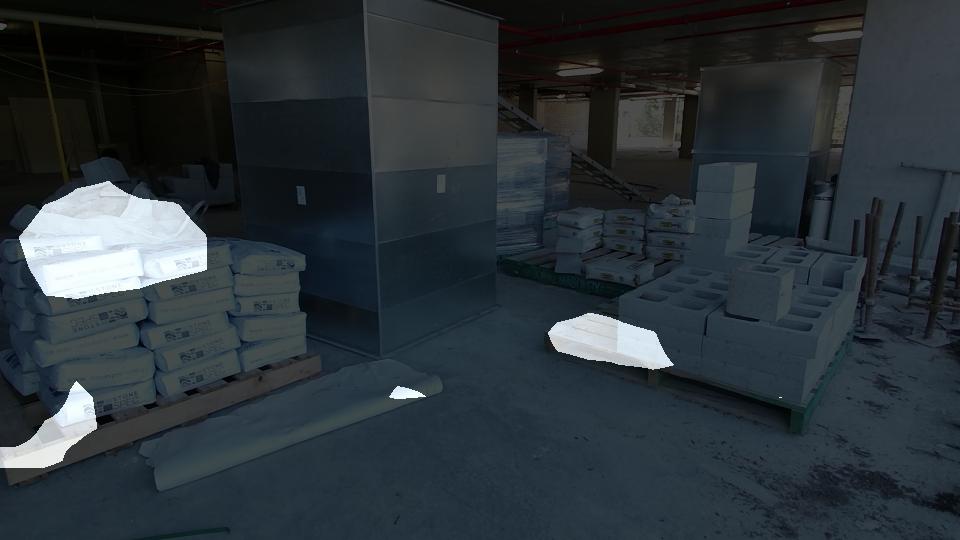}} &
		\subfloat[HHA]{\includegraphics[width=\LM\linewidth,clip]{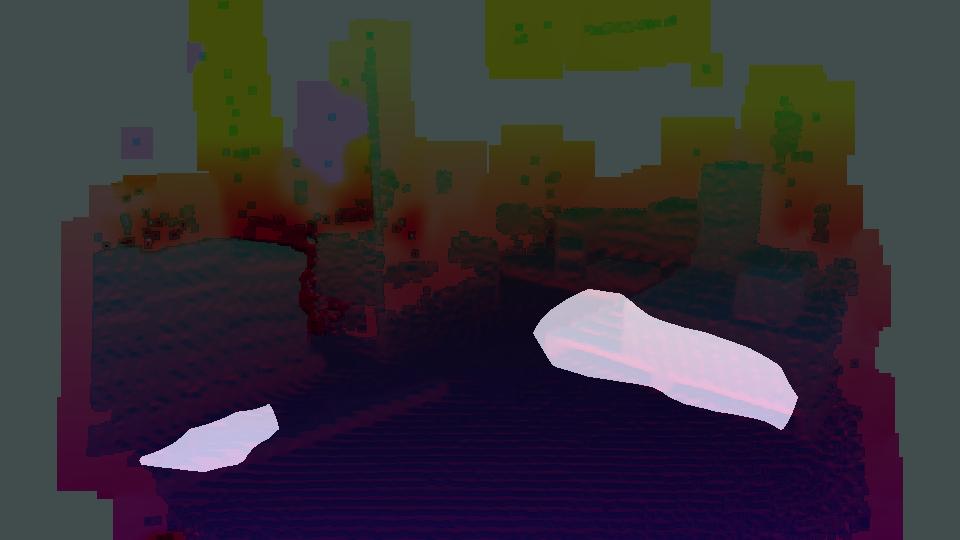}} &
	    \subfloat[Late Prop RGB-HHA]{\includegraphics[width=\LM\linewidth,clip]{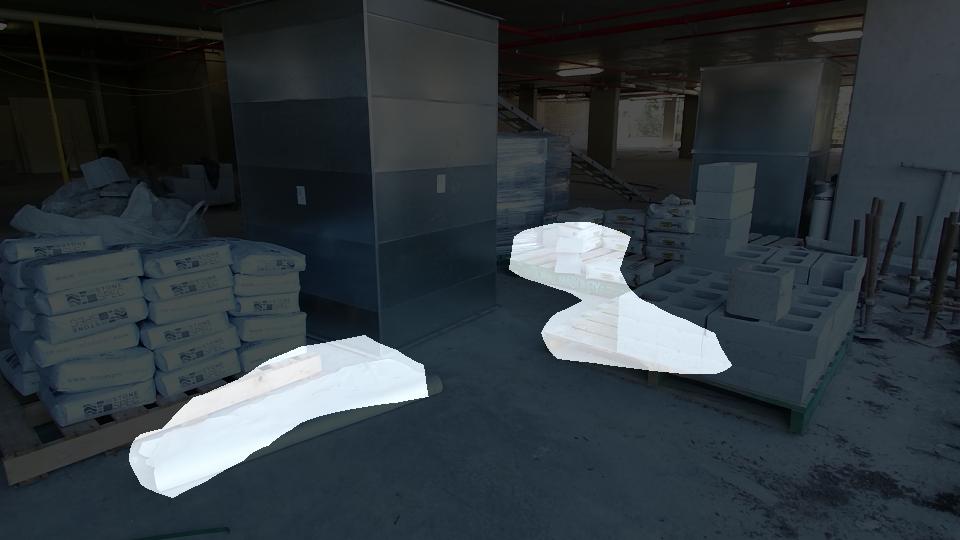}}
	\end{tabular}
	\caption{The Late Proportional Approach (e) leverages information from both modalities RGB (c) and HHA (d) to correctly detect the pallet next to the sliver aluminium pillar as a trip hazard (missed by both non fusion approaches). It also correctly identifies more of the pixels of the large paper roll in the bottom left of the images.}
	\label{fig:late_prop_comparison_74}
	\vspace{-0.5cm}
\end{figure*}

\subsubsection{Late Proportional Fusion Outperforms Other Fusion Approaches}
Although fusing colour and depth information generally tends to improve detection performance, the differences in performance between fusion approaches are significant. Of the fusion approaches, the Late Proportional outperforms the Early, Mid, and Late Overlay Fusion approaches. This can be seen in Figure \ref{fig:PR_curve_fusion} and Table \ref{table:overallresults} where the Late Proportional fusion (RGB-HHA) outperforms all other approaches in a majority of operating points in precision and recall. Figure \ref{fig:Trip_IOU_curve_fusion} shows Late Proportional Fusion achieves the highest trip IOUs for all trip object detection's above 65\%. Despite both using Late Fusion, Late Overlay (RGB-HHA) is outperformed by Late Proportional (RGB-HHA), which is due to how they are trained for fusion. Rather than the fixed, naive fusion of modalities used by Late Overlay Fusion; the Late Proportional fusion approach is \textit{trained} with both modalities allowing it to learn fusion.

Mid Fusion achieves the second best performance in the PR curve (Figure \ref{fig:PR_curve_fusion}) but starts dropping below other approaches in Trip IOU at around 65\% Trip Object Detection (Figure \ref{fig:Trip_IOU_curve_fusion}). Early Fusion sees a similar trend to Mid Fusion with a dropping Trip IOU score in Figure \ref{fig:Trip_IOU_curve_fusion}, and also drops off in precision at higher recall values relative to the other approaches in Figure \ref{fig:PR_curve_fusion}. From an empirical perspective, Early Fusion appears to rely on raw depth or HHA images too much, often failing to detect trip hazards in cases with less depth information such as small trip hazards and hazards out of depth sensing range (see Figure \ref{fig:rgb_vs_hha_res_range}). Mid Fusion shows an improvement over Early Fusion by more effectively leveraging both modalities in regions where depths limited resolution fails to visualise small trip hazards, such as wires. However, compared to Late Proportional Fusion, Mid Fusion fails more frequently to detect more distant hazards commonly out of the depth sensor's range (see Figure \ref{fig:rgb_vs_hha_res_range}(b)).

\subsubsection{HHA Encoded Images are Superior to Raw Depth Images}
The investigation also establishes the improvements from using HHA encoded images over raw depth images as a representation of depth/structural information. This is illustrated for single modality approaches in Figure \ref{fig:PR_curve_nonfusion} where HHA outperforms colour at certain operating points, and when used in fusion (Figures \ref{fig:PR_curve_fusion} and \ref{fig:Trip_IOU_curve_fusion}). Our improvements from using the HHA encoded image over raw depth images matches the observations made by others \cite{Gupta2014a, Long2014}.

\subsubsection{Colour Information is Superior to Depth Information}
We establish in Figure \ref{fig:PR_curve_nonfusion} that the trip detector trained on colour information (using RGB images) outperforms the detectors trained on depth information (raw depth images and HHA images). There are a number of reasons for this, first, FCNs (a type of CNN), are primarily designed to operate with colour image information. Second, there are significantly more labelled colour images available for training than depth/HHA images.
\subsubsection{Sensor Limitations Further Restrict Usefulness of Depth Information}
Another reason colour information is superior to depth information is the limited range and resolution of the depth sensor. Colour sensors have a greater perception range than many depth sensors, such as the Kinect 2 used here, which has a 5 meter depth measurement range. An example of the limited depth perception range can be seen in Figure \ref{fig:rgb_vs_hha_res_range}(b), where a group of distant trip hazards against the wall are missed by the HHA FCN but correctly identified by the RGB. This is because the trip hazards against the wall are not visible in the HHA image as they are out of the depth sensor's range. In addition to the reduced perception range, the limited resolution of many depth sensors restricts the visibility of smaller objects against the background. Figure \ref{fig:rgb_vs_hha_res_range}(a) shows wires on the ground are almost invisible in the HHA image. Here, the RGB trip detector correctly identifies part of the wires as a trip hazard, while the HHA trip detector (unable to even see the wires) fails to detect them. This limited resolution from depth restricts the visibility of small trip hazards (such as wires, hoses and thin pieces of PVC sheeting) to the depth and HHA detectors.

\begin{figure*}
	\centering
	\setlength\tabcolsep{1.0pt}
	\begin{tabular}{ccccc}
		\subfloat[Colour Image]{\includegraphics[width=\LM\linewidth,clip]{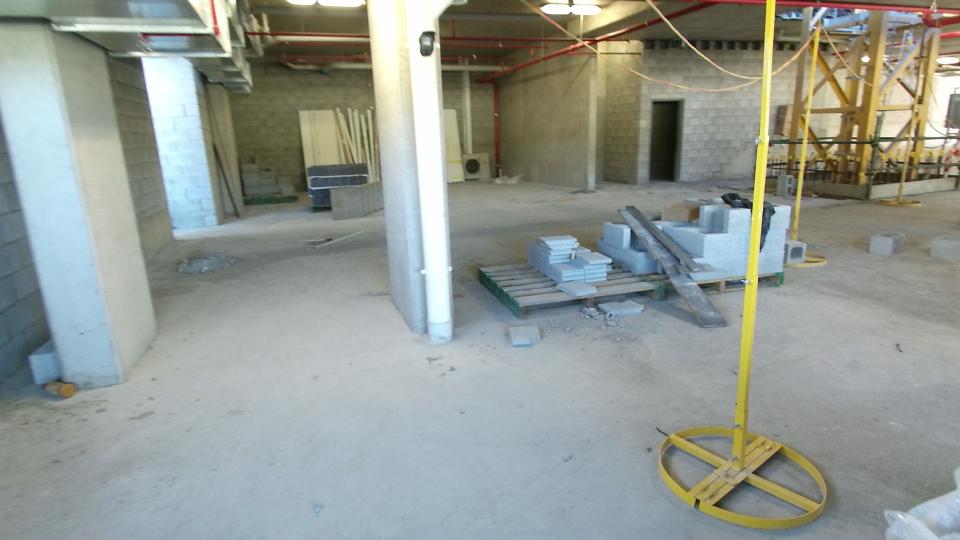}} &
		\subfloat[Ground Truth]{\includegraphics[width=\LM\linewidth,clip]{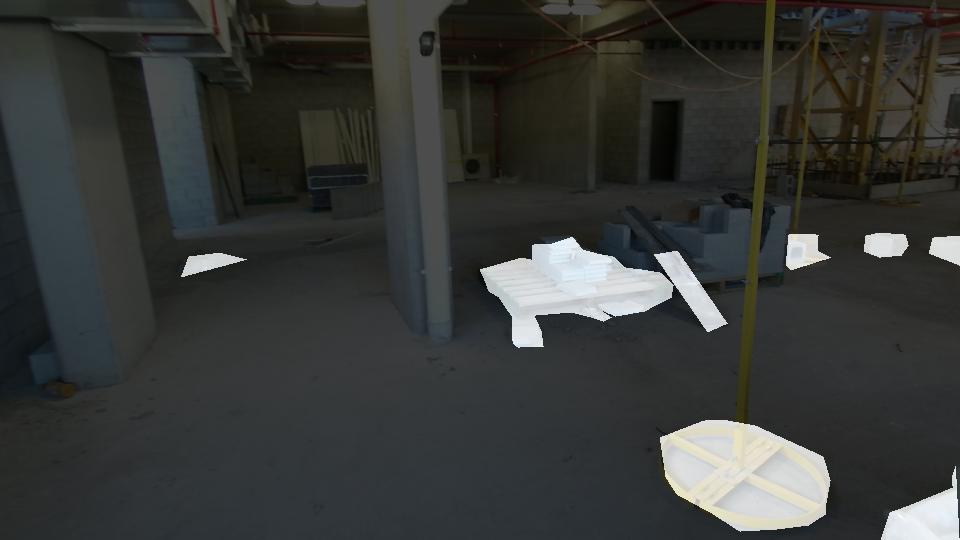}} &
		\subfloat[RGB]{\includegraphics[width=\LM\linewidth,clip]{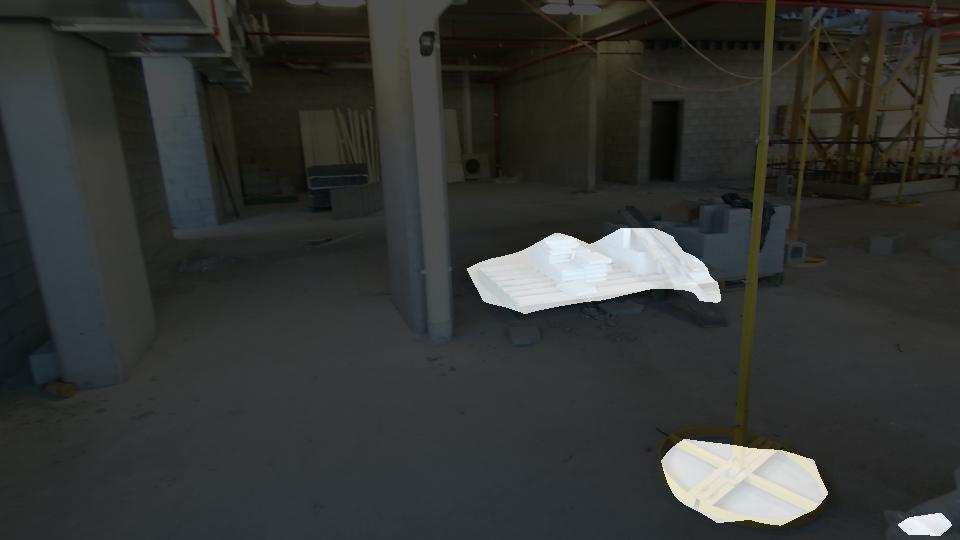}} &
		\subfloat[HHA]{\includegraphics[width=\LM\linewidth,clip]{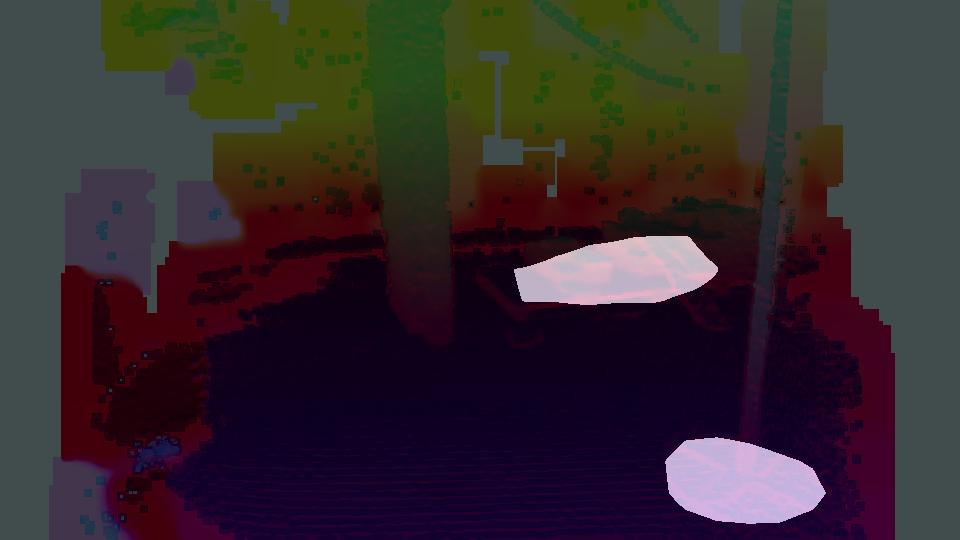}} &
		\subfloat[Late Prop RGB-HHA]{\includegraphics[width=\LM\linewidth,clip]{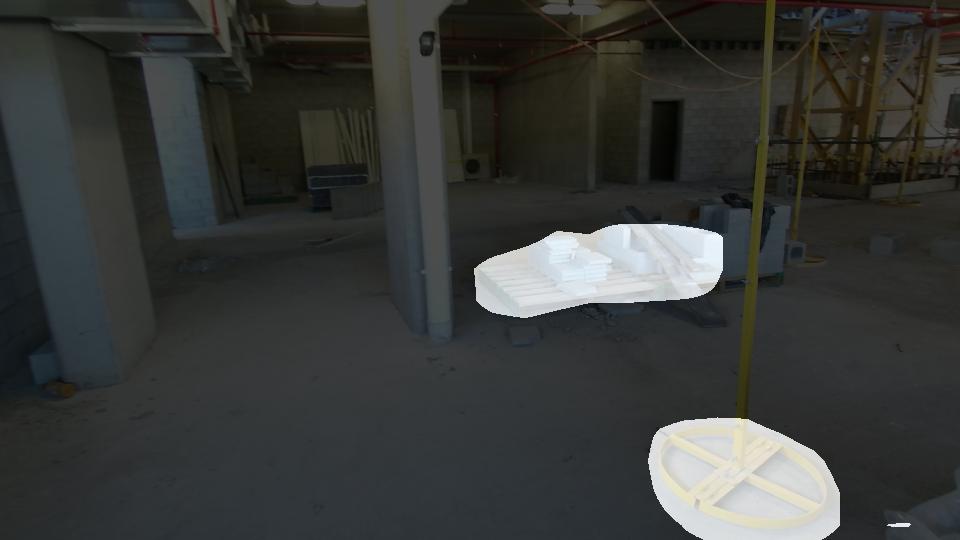}}
	\end{tabular}
	\caption{Adds some explanation for why fusion does not increase performance as significantly as some might expect. Here, two trip hazards, the base of the pole (bottom right) and the pallet (middle) are detected by both colour and HHA. Because of this detection overlap, the fusion of both modalities results in less improvement.}
	\label{fig:late_prop_depth_no_effect_99}
	\vspace{-0.6cm}
\end{figure*}
\subsubsection{Improvements Over Earlier Prototype}
Table \ref{table:tripnet_comp} shows the updated colour trip hazard detector outperforms the earlier prototype colour based trip detector (TripNet \cite{mcmah15}). There are two main reasons for this, first is the different approaches used to locate trip hazards in the image. TripNet simply classifies individual image regions similar to R-CNN \cite{Girshick2014}, restricting spatial information available to the network. While our updated approach uses semantic segmentation which makes the information from the entire image available to the network. Secondly, the performance improvements can also be attributed to the updated network architecture, VGG \cite{Simonyan2014a} (used here), which outperforms AlexNet \cite{Krizhevsky2012} (used in TripNet) in benchmarks \cite{ILSVRC15}.

\section{DISCUSSION}
The approaches and results presented here show that automated trip hazard detection is feasible on real world construction sites using colour and depth sensing.
Through a comprehensive investigation into 11 different colour and depth information fusion and non fusion approaches, it was found that Late Proportional Fusion on RGB and HHA images showed the best performance for trip hazard detection on construction sites. Of the 11 approaches tested, it achieved the highest F1-score and Trip IOU (Table \ref{table:overallresults}) and outperformed other approaches in the majority of operating points (Figures~\ref{fig:PR_curve_nonfusion}, \ref{fig:PR_curve_fusion} and \ref{fig:Trip_IOU_curve_fusion}).

Trip hazards are not the only hazards present on construction sites and on other application domains, however. Future work could involve expanding the multi-modal trip hazard detector to also recognise slip and fall hazards. Performing this work will also reveal how much domain-specific training is required and, whether fine-tuning will be sufficient to adapt trip hazard detection from a construction site to say a ship or oil rig platform.

Operationally, a trip hazard detector could be deployed in at least two different ways. Firstly as an assistive device, that accompanies a safety inspector on their regular site inspections. Secondly, as an automated solution deployed on a ground-based or aerial robotic platform, that performs regular traverses of the site to look for hazards. This research, and the datasets provided, can support future work that will translate to technologies which reduce injuries and fatalities across the wide range of hazardous environments where humans currently work.

{
\bibliographystyle{IEEEtran}
\bibliography{library}
}

\end{document}